\documentclass[12pt]{article}

\usepackage{amsmath}
\usepackage{amssymb}
\usepackage{siunitx}

\usepackage{booktabs}
\usepackage{graphicx}
\usepackage{caption}

\usepackage{geometry}
\usepackage{textcomp}

\usepackage[numbers]{natbib}

\usepackage[colorlinks=true,
            linkcolor=black,
            citecolor=black,
            urlcolor=blue]{hyperref}
\usepackage{orcidlink}
\usepackage{setspace}
\usepackage{authblk}

\begin{document}

\title{A Physics-Informed Neural Operator for Thermal Ranking of Low-Cost Wall Materials in Hot-Dry Climates}

\author[1]{Muhammad Akbar Khan~\orcidlink{0009-0001-7956-0080}\thanks{Corresponding author. Email: \href{mailto:akbar.bsma1337@gmail.com}{akbar.bsma1337@gmail.com}}}
\author[1]{Fahim Raees~\orcidlink{0000-0002-2396-1491}}
\author[1]{Ubaida Fatima~\orcidlink{0000-0003-0372-0858}}
\affil[1]{Department of Mathematics, NED University of Engineering and Technology, Karachi 75270, Pakistan}

\date{}

\maketitle

\begin{abstract}
Identifying cost-effective indigenous building materials that minimise heat penetration
through walls is critical for indoor thermal comfort in low-income rural housing in
hot-dry climates, where summer temperatures routinely exceed 45\,\textdegree C. We present
a two-stage computational framework for thermal ranking of five low-cost indigenous wall
materials: mud brick, clay--straw adobe, lime-stabilised bamboo panel, fired clay
brick, and lime--mud composite. First, a validated Crank--Nicolson finite difference
method (FDM) solves the one-dimensional transient heat equation with Robin boundary
conditions under diurnal solar and outdoor air-temperature forcing, generating 1500
periodic-day solutions across a nine-dimensional parameter space by Latin Hypercube
sampling. Second, a Physics-Informed Neural Operator (PINO) with a Fourier Neural
Operator (FNO) backbone learns the parameter-to-solution operator
$\boldsymbol{\mu} \mapsto T(x,t)$, enforcing both data fidelity and PDE consistency.
The trained PINO attains a relative $L^2$ field error of $5.14\times10^{-4}$ and a
0.201\,K mean absolute error on the peak inner surface temperature, preserving the FDM
material ranking exactly; PINO trained on 150 FDM samples matches a data-only FNO
trained on twice as many, so the physics loss is most valuable when data are scarce.
The periodic-day formulation also yields the ISO~13786 time lag and decrement
factor, reproduced to within 0.99\,h and 0.010. At nominal hot-dry summer
conditions, clay--straw adobe achieves the best cost--performance index among widely
available materials. A climate sweep, confirmed by FDM spot checks, reveals a regime
boundary: under sub-ambient outdoor conditions the ranking inverts to conductive
fired clay brick, delineating heat-exclusion and heat-rejection regimes.
The framework supports evidence-based material selection for post-flood reconstruction
in hot-dry regions.
\end{abstract}

\section{Introduction}
\label{sec:intro}

Rural communities in Sindh, Pakistan, face acute thermal stress
during summer months, with outdoor air temperatures regularly
exceeding \SI{45}{\celsius} and peak solar irradiance reaching
\SI{900}{\watt\per\metre\squared}~\cite{NasaSSE}.
The building envelope, particularly the external wall, is
the primary barrier between outdoor heat and indoor living
conditions, and its thermal performance directly determines
whether a dwelling remains habitable without mechanical
cooling~\cite{Jannat2020,Mehmood2022}; instrumented field
measurements in low-income South Asian households confirm
that roof and wall material alone can produce indoor--outdoor
air temperature differentials of several degrees Celsius in
either direction, with indoor air up to 3--4\textdegree{}C
warmer than outdoor in some material configurations and
measurably cooler than outdoor at peak daytime hours in
others, even in the absence of any cooling
device~\cite{Tasgaonkar2022}.
For the rural poor, mechanical cooling is either unaffordable
or unavailable~\cite{Mehmood2022}: the 2022 Sindh floods damaged up to two million homes
in the province~\cite{Ebrahim2024,Iqbal2025}, accelerating demand for
rapidly constructible, low-cost housing in which passive
thermal performance is the only available mechanism for
indoor comfort.
Earthen wall assemblies have been shown to provide substantially
lower diurnal temperature swings and greater passive thermal
stability than conventional construction across hot-dry and
arid climates, even without mechanical cooling~\cite{BenAlon2023}.
Indigenous earthen materials (mud brick, clay--straw adobe,
lime-stabilised bamboo panel, fired clay brick, and lime--mud
composite) are locally available
throughout rural Sindh, require no imported components, and
have been used in vernacular construction in the
region~\cite{Ebrahim2024,Iqbal2025,Abbasi2025}; a recent meta-analysis
of over 500 studies confirms that earthen construction
techniques including adobe and compressed earth blocks remain
active research areas globally, particularly in hot and
developing-country contexts~\cite{MoraRuiz2025}.
However, their transient thermal performance under realistic
Sindh summer loading has not been systematically quantified
across the full range of material and climate parameters
relevant to post-flood reconstruction practice.

Transient thermal simulation of building envelope materials has
been widely studied using finite difference and finite element
methods.
Jannat et al.~\cite{Jannat2020} performed dynamic thermal
simulations of four base wall materials across thirteen
composite wall configurations in a tropical climate,
demonstrating that thermal mass, not conductivity
alone, governs peak inner surface temperature under daytime
solar loading.
For earthen and clay-based materials specifically, Oti et
al.~\cite{Oti2010} measured design thermal values for unfired
clay bricks; in-situ monitoring of an unfired clay brick
building confirmed thermal lags of approximately 10~h and
indoor temperature swings below 3\,\textdegree C even with
16\,\textdegree C outdoor variation~\cite{ElFgaier2015};
and a comprehensive review of waste-modified unfired earth
bricks reports thermal conductivity spanning
0.14--1.0~W\,m$^{-1}$K$^{-1}$ depending on composition and
fibre content~\cite{Lachheb2023}.
Laaroussi et al.~\cite{Laaroussi2014}
characterised the thermal conductivity of fired clay brick
using multiple experimental methods, reporting a value of
approximately $0.35$~W\,m$^{-1}$K$^{-1}$.
The thermal properties of clay--straw composites were
investigated by El Azhary et al.~\cite{ElAzhary2017}, who
reported a 48\% reduction in thermal conductivity at a 5\%
straw mass fraction relative to pure clay.
Moisture content raises the effective thermal conductivity of
porous wall materials in a manner well described by a linear
mixing rule~\cite{Wei2019}, a behaviour documented broadly
across raw earth construction types including adobe, rammed
earth, and compressed earth blocks~\cite{Giada2019}, and
this effect must be accounted for in parametric studies
that span a range of initial moisture states.
On the operator learning side, neural operators more broadly
have demonstrated mesh-independent, differentiable approximations
for parametric PDEs with speedups of four to five orders of
magnitude over traditional solvers~\cite{Azizzadenesheli2024},
and Fourier Neural Operators (FNO)~\cite{Li2021FNO} have
demonstrated mesh-independent operator learning for parametric
PDEs, and the Physics-Informed Neural Operator
(PINO)~\cite{Li2021PINO} extends FNO training with a PDE
residual loss that enforces physical consistency without
additional labelled data.
PINO has been demonstrated on Burgers', Darcy, and
Navier--Stokes flow problems~\cite{Li2021PINO}, and physics-informed neural networks~\cite{Raissi2019}
more broadly have been used to solve convection problems with
unknown thermal boundary conditions, two-phase Stefan problems,
and industrial heat-sink design~\cite{Cai2021}, as
well as heat conduction in porous media with no labelled training
data~\cite{Xu2023}; operator-learning approaches have also been
applied directly to parametric heat conduction with variable
source terms~\cite{Koric2023}. A recent comprehensive review
surveys PINN applications across heat-transfer-dominated
multiphysics systems~\cite{Zhao2025}. However, the application
of physics-informed operator learning to the parametric thermal
analysis of building envelope materials has not been explored.

No existing study has applied a neural operator to
the parametric thermal analysis of indigenous building materials
in a developing-country context, nor to the specific conditions
of rural Sindh.
Prior computational comparisons of indigenous wall materials
have relied on single-configuration simulations at
fixed parameter values~\cite{Jannat2020,Oti2010,ElAzhary2017,Ali2025},
without exploring the nine-dimensional space of material
properties, wall geometry, moisture content, and climate
variables that characterises the practical variability of
rural Sindh construction.
Furthermore, no prior work has combined transient thermal
simulation with a cost--performance index expressed in local
currency, which is essential for evidence-based material
selection in low-income rural communities where procurement
decisions are made in Pakistani rupees (PKR), not normalised
thermal resistance values.

The present study addresses these gaps with the following
contributions.
First, a Crank--Nicolson FDM solver with diurnal, time-varying
boundary forcing is developed, validated via
the method of manufactured solutions, a Robin boundary condition
zero-drift test, a space--time convergence study of the
time-varying problem against a fine reference, and periodicity
and initial-condition-independence checks, and used to generate 1500
high-fidelity periodic-day temperature field solutions across a
nine-dimensional parameter space via Latin Hypercube Sampling.
Second, a PINO with an FNO backbone is trained on
this dataset, enforcing both data fidelity and PDE consistency,
and is benchmarked against a data-only FNO baseline.
Third, the trained PINO is used to rank five indigenous
Sindh wall materials by peak inner surface temperature,
complemented by the ISO~13786 dynamic metrics of thermal time
lag and decrement factor that the periodic-day formulation
makes available, and to construct a cost--performance index
that combines simulated thermal performance with indicative
local material costs in PKR/m$^2$.
Fourth, a data-efficiency study quantifies where the physics
loss pays off, and a climate sweep of the trained operator,
confirmed by FDM ground truth, maps a physically meaningful
heat-exclusion/heat-rejection regime boundary in the climate
design space.
The remainder of this paper is organised as follows.
Section~\ref{sec:phy} presents the physical model, governing heat equation, and effective thermal conductivity.
Section~\ref{sec:bcs} details the solar forcing model, the boundary conditions, and the initial conditions.
Section~\ref{sec:params} defines the parametric study and
candidate materials.
Section~\ref{sec:method} describes the FDM data generation and
PINO training.
Section~\ref{sec:results} reports the validation and accuracy results, including the material cost--performance assessment.
Section~\ref{sec:discussion} discusses the results and study
limitations, and Section~\ref{sec:conclusion} concludes.
\section{Physical Model}
\label{sec:phy}

\begin{figure}[htbp]
\centering
\includegraphics[width=\textwidth]{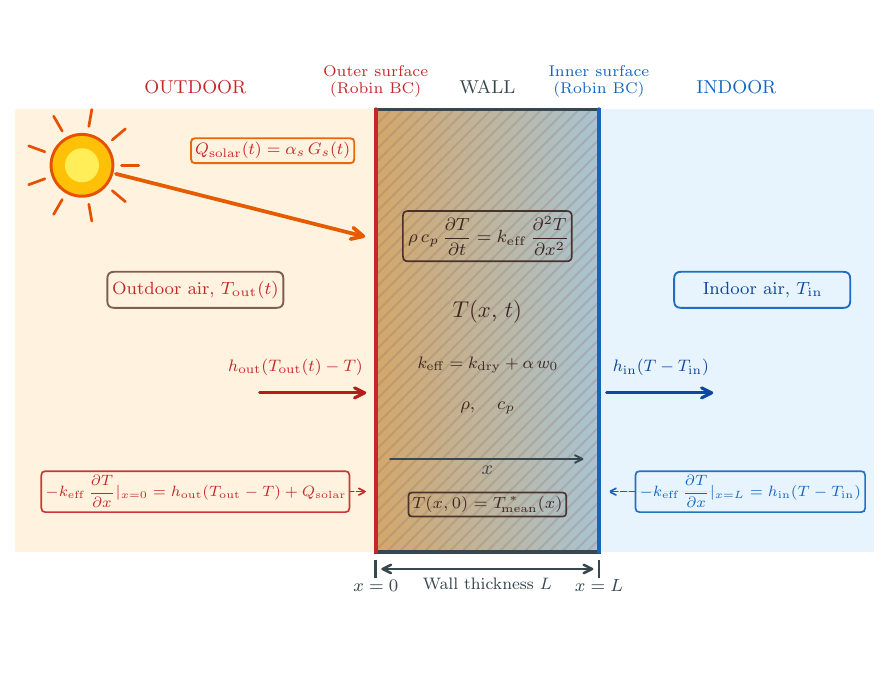}
\caption{Schematic of the one-dimensional transient heat conduction
problem. The outer surface ($x=0$) is exposed to outdoor convection
and absorbed solar radiation under a Robin boundary condition; the
inner surface ($x=L$) exchanges heat with indoor air under a second
Robin boundary condition. The wall is characterised by effective
thermal conductivity $k_{\mathrm{eff}}$, density $\rho$, and specific
heat capacity $c_p$, initialised at the steady state of the
mean-forcing initial condition, Eq.~\eqref{eq:ic_T}; the
diurnal forcing model is shown in Fig.~\ref{fig:forcing}.}
\label{fig:wall_schematic}
\end{figure}

Figure~\ref{fig:wall_schematic} illustrates the one-dimensional
wall domain, boundary conditions, and initial condition considered
in this study.
One partial differential equation governs heat transfer through
the wall material.
The initial moisture content of the wall affects heat transfer
through the effective thermal conductivity $k_{\mathrm{eff}}$,
defined in Section~\ref{sec:keff}.

\subsection{Heat Equation}
\label{sec:heat}

Energy conservation in the wall gives:
\begin{equation}
    \rho\, c_p \frac{\partial T}{\partial t}
    = \frac{\partial}{\partial x}
      \!\left(k_{\mathrm{eff}}\,
      \frac{\partial T}{\partial x}\right)
    \label{eq:heat}
\end{equation}

\noindent
where $t$ (\si{\second}) is time, $x$ (\si{\metre}) is the
spatial coordinate across the wall thickness, $\rho$
(\si{\kilogram\per\cubic\metre}) is the wall material density,
$c_p$ (\si{\joule\per\kilogram\per\kelvin}) is the specific heat
capacity, $T$ (\si{\kelvin}) is temperature, and
$k_{\mathrm{eff}}$ (\si{\watt\per\metre\per\kelvin}) is the
effective thermal conductivity defined in Section~\ref{sec:keff}.
Solar radiation enters the problem as a surface heat flux
through the outer boundary condition \eqref{eq:bc_out};
it does not appear as a source term in \eqref{eq:heat}.

Latent heat of vaporisation is neglected throughout.
The effect of moisture on heat transfer enters solely through
$k_{\mathrm{eff}}$ \eqref{eq:keff}.
This is a conservative simplification: for earthen materials
(mud brick, clay--straw adobe) with elevated
moisture content, evaporative cooling would reduce the inner
surface temperature in practice~\cite{BenAlon2023}, so the
present model gives an upper bound on $J$ for those materials.

\subsection{Moisture Content and Effective Thermal Conductivity}
\label{sec:keff}

Moisture raises the thermal conductivity of porous wall
materials linearly \cite{Wei2019}:
\begin{equation}
    k_{\mathrm{eff}} = k_{\mathrm{dry}} + \alpha\, w_0
    \label{eq:keff}
\end{equation}
where $k_{\mathrm{dry}}$ (\si{\watt\per\metre\per\kelvin})
is the dry-state thermal conductivity, $\alpha$
(\si{\watt\metre\squared\per\kilogram\per\kelvin}) is the
moisture--conductivity coupling coefficient, and $w_0$
(\si{\kilogram\per\cubic\metre}) is the initial volumetric
moisture content of the wall.
This linear form follows established precedent in the
building-physics literature, where it is among the most
commonly used empirical relations between moisture content
and effective thermal conductivity~\cite{Liu2016}.
Experimental data show this linear form to be most accurate
at small-to-moderate moisture contents; a sub-linear
power-law fit improves accuracy at higher moisture contents
and larger pore sizes~\cite{Liu2016}.
The swept range of $w_0$ adopted here
(0--\SI{30}{\kilogram\per\cubic\metre}, Table~\ref{tab:params})
remains modest relative to the near-saturation moisture
contents at which this nonlinearity becomes
pronounced~\cite{Liu2016}, supporting the linear approximation
as adequate across the present parameter range.

Moisture is treated as spatially uniform and constant
throughout the simulation at its initial value $w_0$.
This is justified by the diffusion timescale argument in
Section~\ref{sec:ics}: the characteristic time for moisture
to redistribute across the wall ($\tau_m \approx 26$~days,
Eq.~\eqref{eq:tau_m}) far exceeds the 120-hour simulation
window.
$k_{\mathrm{eff}}$ is therefore computed once from
\eqref{eq:keff} before the time loop and remains constant
throughout each simulation run.
Both $w_0$ and $\alpha$ are retained as free parameters in
$\boldsymbol{\mu}$ (Table~\ref{tab:params}), so their
combined effect on thermal performance is fully explored
by the parametric sweep.

The moisture--conductivity coupling coefficient $\alpha$ is
not tabulated in the open literature for the specific
indigenous materials studied here.
Nominal values in Table~\ref{tab:materials} are assigned by
the authors at the same order of magnitude as the coefficient
reported by Wei et al.\ \cite{Wei2019} for a moisture-sensitive
cementitious wall material; the effect of uncertainty in
$\alpha$ across the full parametric range $[0.001, 0.010]$
(Table~\ref{tab:params}) is explored through the parametric
sweep itself.

More detailed heat, air and moisture (HAM) formulations
employ a coupled moisture diffusion PDE with
capillary-pressure or vapour-pressure driven
transport \cite{Indekeu2022}.
The present model captures only the effect of static
moisture distribution on conductivity via \eqref{eq:keff};
it neglects the additional conductivity contribution from
active moisture migration and phase change, which has been
shown to become significant under cooling conditions with
large indoor/outdoor water-vapour partial-pressure
differences~\cite{Wang2019}.
Rural Sindh's hot-dry summer climate is characterised by low
ambient humidity and correspondingly small vapour-pressure
gradients relative to the humid climates in which this effect
has been quantified~\cite{Wang2019}, which limits the
magnitude of the neglected term in the present context.
Such formulations become relevant for multi-day simulations
in which moisture redistribution is dynamically significant;
that extension is identified as future work in
Section~\ref{sec:conclusion}.
\section{Solar Forcing, Boundary Conditions, and Initial Conditions}
\label{sec:bcs}

This section specifies the three ingredients that close the
physical model of Section~\ref{sec:phy}: the diurnal solar
heat source absorbed at the outer surface, the Robin boundary
conditions coupling both wall faces to their environments, and
the initial condition from which the periodic quasi-steady
state is reached.

\subsection{Solar Heat Source}
\label{sec:solar}

\begin{figure}[htbp]
\centering
\includegraphics[width=0.75\textwidth]{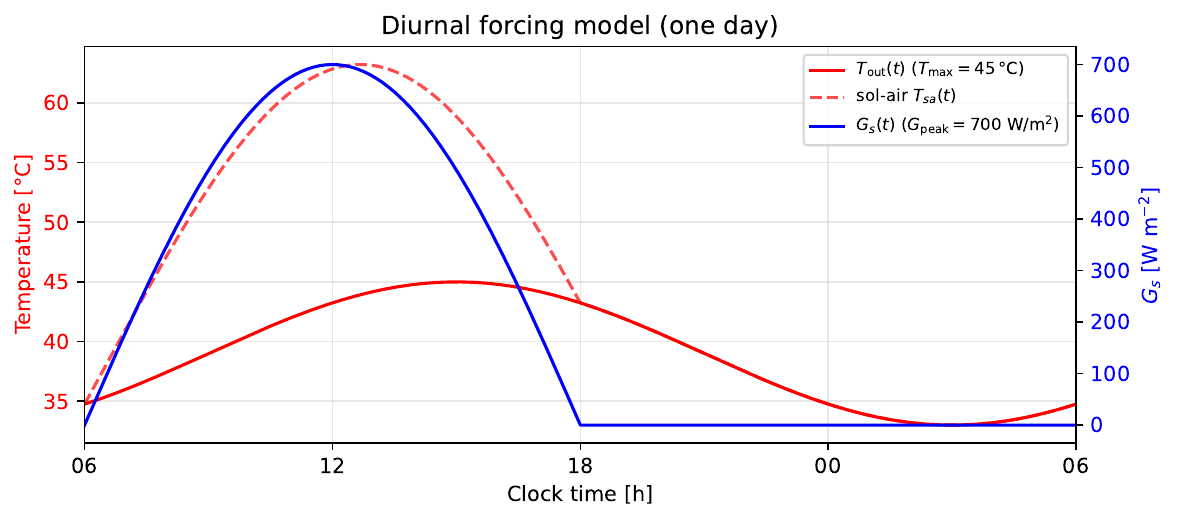}
\caption{Diurnal forcing model over one day at nominal Sindh
         conditions ($T_{\mathrm{out,max}} = \SI{45}{\celsius}$,
         $G_{s,\mathrm{peak}} = \SI{700}{\watt\per\metre\squared}$):
         outdoor air temperature $T_{\mathrm{out}}(t)$
         \eqref{eq:tout_profile} with maximum at 15:00, half-sine
         solar irradiance $G_s(t)$ \eqref{eq:gs_profile} spanning
         06:00--18:00, and the resulting sol-air temperature
         $T_{sa}(t)$ \eqref{eq:solair}.}
\label{fig:forcing}
\end{figure}

Solar radiation is absorbed at the outer wall surface and
enters the problem as a surface heat flux through the outer
Robin boundary condition \eqref{eq:bc_out} of
Section~\ref{sec:bc_heat}.
The incident irradiance follows a half-sine clear-sky diurnal
profile spanning the 12-hour daylight window, with $t = 0$
defined at sunrise (06:00 local time) and
$\tau_d = t \bmod \SI{24}{\hour}$ the time of day:
\begin{equation}
    G_s(t)
    =
    \begin{cases}
        G_{s,\mathrm{peak}}\,
        \sin\!\left(\dfrac{\pi\,\tau_d}{t_{\mathrm{day}}}\right),
        & 0 \le \tau_d < t_{\mathrm{day}}, \\[6pt]
        0, & t_{\mathrm{day}} \le \tau_d < \SI{24}{\hour},
    \end{cases}
    \label{eq:gs_profile}
\end{equation}
where $t_{\mathrm{day}} = \SI{12}{\hour}$ and
$G_{s,\mathrm{peak}}$
(\si{\watt\per\metre\squared}) is the solar-noon peak
irradiance, a parametric input (Table~\ref{tab:params}).
The absorbed flux entering the outer Robin condition
\eqref{eq:bc_out} is:
\begin{equation}
    Q_{\mathrm{solar}}(t)
    = \alpha_s\, G_s(t)
    \label{eq:qsolar}
\end{equation}
where $\alpha_s = 0.7$ is the solar absorptivity of the outer
wall surface, consistent with the default value for unplastered
opaque surfaces in standard building simulation practice~\cite{EnergyPlus}
and within the range measured for earthen wall materials.
For representative Sindh conditions
($G_{s,\mathrm{peak}} = \SI{700}{\watt\per\metre\squared}$,
$\alpha_s = 0.7$) the solar-noon absorbed flux is
$Q_{\mathrm{solar}} = \SI{490}{\watt\per\metre\squared}$.
Figure~\ref{fig:forcing} illustrates the forcing model over
one day.
\subsection{Boundary Conditions for the Heat Equation}
\label{sec:bc_heat}

\noindent
\textbf{Outer surface $x = 0$ (Robin boundary condition, BC).}
The outer face is exposed to solar radiation and outdoor
convection:
\begin{equation}
    -k_{\mathrm{eff}}\,
    \frac{\partial T}{\partial x}\bigg|_{x=0}
    = h_{\mathrm{out}}\bigl(T_{\mathrm{out}}(t) - T\bigr)
      + Q_{\mathrm{solar}}(t)
    \label{eq:bc_out}
\end{equation}
where $h_{\mathrm{out}} = \SI{25}{\watt\per\metre\squared\per\kelvin}$
is the outdoor combined convective and radiative heat transfer
coefficient, taken from the surface resistance values in ISO~6946~\cite{ISO6946}
(a steady-state standard; ISO~13786~\cite{ISO13786} governs dynamic thermal
characterisation of building components) and
$T_{\mathrm{out}}(t)$ is the diurnally varying outdoor air
temperature; the absorbed solar flux $Q_{\mathrm{solar}}(t)$
entering \eqref{eq:bc_out} is defined in
Section~\ref{sec:solar}.
Consistent with the sinusoidal air-temperature models used in
periodic building thermal analysis~\cite{ISO13786}, $T_{\mathrm{out}}(t)$
follows:
\begin{equation}
    T_{\mathrm{out}}(t)
    = \bigl(T_{\mathrm{out,max}} - A\bigr)
    + A\,\sin\!\left(
        \frac{2\pi\,(\tau_d - t_{\varphi})}{\SI{24}{\hour}}
      \right),
    \qquad
    A = \tfrac{1}{2}\,\Delta T_{\mathrm{diurnal}},
    \label{eq:tout_profile}
\end{equation}
where $T_{\mathrm{out,max}}$ (\si{\celsius}) is the daily
maximum air temperature, a parametric input
(Table~\ref{tab:params}); the total diurnal swing is fixed at
$\Delta T_{\mathrm{diurnal}} = \SI{12}{\kelvin}$, a conservative
value within the range observed in upper-Sindh summer
conditions, where NASA POWER daily temperature ranges at
Sukkur span approximately 7--23~K with a May--June median
near 16~K~\cite{NasaSSE}; a smaller swing understates
inter-material contrast, so the ranking results are
conservative with respect to this choice; and the phase shift
$t_{\varphi} = \SI{3}{\hour}$ places the air-temperature
maximum at 15:00 and the minimum at 03:00, lagging solar noon
by approximately three hours, consistent with measured diurnal
profiles at South Asian weather stations, including semi-arid
Faisalabad, Pakistan~\cite{Tasgaonkar2022}.
The sol-air temperature
\begin{equation}
    T_{sa}(t)
    = T_{\mathrm{out}}(t)
    + \frac{\alpha_s\, G_s(t)}{h_{\mathrm{out}}}
    \label{eq:solair}
\end{equation}
combines both forcing components into the single equivalent
driving temperature used as the reference signal for the
dynamic metrics of Section~\ref{sec:params}.

\medskip
\noindent
\textbf{Inner surface $x = L$ (Robin BC).}
The inner face loses heat to the indoor air:
\begin{equation}
    -k_{\mathrm{eff}}\,
    \frac{\partial T}{\partial x}\bigg|_{x=L}
    = h_{\mathrm{in}}\bigl(T - T_{\mathrm{in}}\bigr)
    \label{eq:bc_in}
\end{equation}
where $h_{\mathrm{in}} = \SI{7.69}{\watt\per\metre\squared\per\kelvin}$
is the indoor combined (convective + radiative) heat transfer coefficient,
consistent with the internal surface resistance $R_{si} = 0.13$~\si{\metre\squared\kelvin\per\watt}
specified in ISO~6946~\cite{ISO6946}, and
$T_{\mathrm{in}}$ is the indoor air temperature,
a parametric input listed in Table~\ref{tab:params},
covering the range from passively cooled to fully
unconditioned rural housing in Sindh.

\medskip
\noindent
\textbf{Analytical steady state under constant forcing.}
For constant $T_{\mathrm{out}}$ and $Q_{\mathrm{solar}}$, the
Robin system \eqref{eq:bc_out}--\eqref{eq:bc_in} admits an
exact steady state, used both for the mean-forcing initial
condition (Section~\ref{sec:ics}) and for the solver
validation (Section~\ref{sec:fdm_val}).
The uniform steady-state heat flux through the wall is:
\begin{equation}
    q^*
    = \frac{T_{\mathrm{out}}
            + Q_{\mathrm{solar}}/h_{\mathrm{out}}
            - T_{\mathrm{in}}}
           {1/h_{\mathrm{out}} + L/k + 1/h_{\mathrm{in}}}
    \label{eq:q_robin}
\end{equation}
and the corresponding temperature profile is:
\begin{equation}
    T^*(x)
    = T_{\mathrm{out}}
    + \frac{Q_{\mathrm{solar}} - q^*}{h_{\mathrm{out}}}
    - \frac{q^*}{k}\,x
    \label{eq:T_robin}
\end{equation}
which can be verified to satisfy both Robin conditions
\eqref{eq:bc_out}--\eqref{eq:bc_in} exactly.

\subsection{Initial Conditions}
\label{sec:ics}

Because the analysis targets the periodic quasi-steady
response of the wall to the diurnal forcing
\eqref{eq:gs_profile} and \eqref{eq:tout_profile}, rather
than any particular start-up transient, the initial
condition serves only to initialise the spin-up and does not
influence the reported results.
To accelerate spin-up, the solver is initialised at the exact
Robin steady state \eqref{eq:T_robin} evaluated under the
\emph{time-mean} forcing,
\begin{equation}
    T(x,\, 0) = T^*_{\mathrm{mean}}(x),
    \qquad
    \bar{T}_{\mathrm{out}} = T_{\mathrm{out,max}} - A,
    \quad
    \bar{Q}_{\mathrm{solar}} = \frac{\alpha_s\, G_{s,\mathrm{peak}}}{\pi},
    \label{eq:ic_T}
\end{equation}
which, for this linear problem, coincides with the time-mean of
the periodic solution, so that only the zero-mean periodic
ripple remains to be established during spin-up.
The initial moisture content $w_0$ is a free parameter
(Table~\ref{tab:params}); together with $\alpha$ it
determines $k_{\mathrm{eff}}$ via \eqref{eq:keff}.

\medskip\noindent\textbf{Simulation protocol.}
The FDM solver integrates $N_{\mathrm{days}} = 5$ identical
forcing days ($t_{\mathrm{end}} = \SI{120}{\hour}$) and
extracts the final 24-hour day, by which point the wall has
reached its periodic quasi-steady state.
Periodicity is verified per sample by the metric
$\max_t |T(L,t)_{\mathrm{day\,4}} - T(L,t)_{\mathrm{day\,5}}|$,
which does not exceed \SI{0.043}{\kelvin} across the entire
1500-sample sweep (Section~\ref{sec:fdm_val}); the final day
is likewise verified to be independent of the choice of initial
condition.
All PINO training and evaluation uses this final-day solution
on a 24-hour horizon.

\medskip\noindent\textbf{Moisture redistribution timescale.}
The characteristic equilibration time for moisture across a
wall of thickness $L$ under Fickian diffusion with moisture
diffusivity $D_w$ (\si{\square\metre\per\second}) is
$\tau_m = L^2/D_w$.
A representative value $D_w = \SI{1e-8}{\square\metre\per\second}$,
consistent with the order of magnitude of moisture diffusivity
measured at the low moisture contents considered here for
rammed earth~\cite{Indekeu2022} and for clay-based porous
building materials~\cite{Koci2018}, is used here; both sources
report diffusivity rising by several orders of magnitude
toward capillary saturation, outside the range explored in
this study.
For the minimum wall thickness in Table~\ref{tab:params}:
\begin{equation}
    \tau_m
    \;=\;
    \frac{(0.15)^2}{10^{-8}}
    \;=\; \SI{2.25e6}{\second}
    \;\approx\; 26~\text{days}
    \label{eq:tau_m}
\end{equation}
Since $t_{\mathrm{end}} = \SI{120}{\hour} \ll \tau_m \approx
\SI{624}{\hour}$, moisture redistribution remains small during
any simulation run:
$w(x,t) \approx w_0$ throughout $[0,\,t_{\mathrm{end}}]$ for
all five candidate materials.
The static treatment of $w_0$ in \eqref{eq:keff} is therefore
physically justified for the present hot-dry peak-summer
analysis window.

\medskip\noindent\textbf{Thermal equilibration timescale.}
The characteristic thermal equilibration time of the wall is
\begin{equation}
    \tau_T
    \;=\;
    \frac{\rho\,c_p\,L^2}{k}
    \;=\;
    \frac{L^2}{\alpha_{\mathrm{th}}}
    \label{eq:tau_T}
\end{equation}
where $\alpha_{\mathrm{th}} = k/(\rho c_p)$
(\si{\metre\squared\per\second}) is the thermal diffusivity.
At the midpoint wall thickness $L = \SI{0.25}{\metre}$,
$\tau_T$ ranges from \SI{40.4}{\hour} (fired clay brick,
$\alpha_{\mathrm{th}} = \SI{4.30e-7}{\metre\squared\per\second}$)
to \SI{112}{\hour} (lime-stabilised bamboo panel,
$\alpha_{\mathrm{th}} = \SI{1.54e-7}{\metre\squared\per\second}$)
across the five candidate materials.
Under periodic forcing, $\tau_T$ governs the decay of the
initial transient towards the periodic attractor: the slowest
spatial mode decays approximately as
$\exp(-\pi^2 t / \tau_T)$, so the five-day spin-up window
suppresses the initial transient by several orders of magnitude
even for the most sluggish wall in the parameter space.
This is confirmed numerically by the periodicity and
initial-condition-independence checks in
Section~\ref{sec:fdm_val}; the extracted final day therefore
represents the periodic quasi-steady response of the wall to
the diurnal Sindh forcing, free of start-up artefacts.
\section{Parametric Study}
\label{sec:params}

The governing equation and boundary conditions remain fixed.
Only the material and climate parameters vary.
The full parameter vector is:
\begin{equation}
    \boldsymbol{\mu}
    = \bigl(
        k_{\mathrm{dry}},\;
        \rho,\;
        c_p,\;
        \alpha,\;
        L,\;
        w_0,\;
        T_{\mathrm{out,max}},\;
        T_{\mathrm{in}},\;
        G_{s,\mathrm{peak}}
      \bigr)
    \label{eq:params}
\end{equation}

Table~\ref{tab:params} summarises all parameters and their
ranges, chosen to span the realistic spectrum of indigenous
Sindh building materials and Sindh climate conditions
\cite{NasaSSE}.

\begin{table}[htbp]
\centering
\caption{Parametric sweep variables for the Sindh housing study.}
\label{tab:params}
\begin{tabular}{lllll}
\toprule
Symbol & Parameter & Unit & Range & Group \\
\midrule
$k_{\mathrm{dry}}$ & Dry thermal conductivity
    & \si{\watt\per\metre\per\kelvin}
    & 0.20--1.50  & Material \\
$\rho$ & Density
    & \si{\kilogram\per\cubic\metre}
    & 800--2200   & Material \\
$c_p$ & Specific heat capacity
    & \si{\joule\per\kilogram\per\kelvin}
    & 650--2000   & Material \\
$\alpha$ & Moisture--conductivity coefficient
    & \si{\watt\metre\squared\per\kilogram\per\kelvin}
    & 0.001--0.010 & Material \\
$L$ & Wall thickness
    & \si{\metre}
    & 0.15--0.45  & Geometry \\
$w_0$ & Initial moisture content
    & \si{\kilogram\per\cubic\metre}
    & 0--30       & Moisture \\
$T_{\mathrm{out,max}}$ & Daily-maximum outdoor air temperature \eqref{eq:tout_profile}
    & \si{\celsius}
    & 25--48      & Climate  \\
$T_{\mathrm{in}}$ & Indoor air temperature
    & \si{\celsius}
    & 25--45      & Climate  \\
$G_{s,\mathrm{peak}}$ & Peak (solar-noon) irradiance \eqref{eq:gs_profile}
    & \si{\watt\per\metre\squared}
    & 200--900    & Climate  \\
\bottomrule
\end{tabular}
\end{table}

\noindent
Temperature parameters $T_{\mathrm{out,max}}$ and $T_{\mathrm{in}}$ are
listed in \si{\celsius} for readability; they are converted to
\si{\kelvin} before use in the governing equations.

\noindent
\textbf{Quantity of interest (QoI).}
The QoI is the temperature at the inner wall surface as a
function of time over the final, periodic quasi-steady day
(Section~\ref{sec:ics}):
\begin{equation}
    J(t;\,\boldsymbol{\mu})
    = T\!\left(L,\, t;\,\boldsymbol{\mu}\right),
    \qquad t \in [0,\, \SI{24}{\hour}]
    \label{eq:qoi}
\end{equation}
The scalar quantity of interest used for material ranking and the
cost--performance index (Section~\ref{sec:cost}) is the peak inner
surface temperature over the final day:
\begin{equation}
    J(\boldsymbol{\mu})
    = \max_{t\,\in\,[0,\,\SI{24}{\hour}]}
      J\!\left(t;\,\boldsymbol{\mu}\right)
    \label{eq:qoi_peak}
\end{equation}
A material is preferred if it produces a lower
$J(\boldsymbol{\mu})$ \eqref{eq:qoi_peak},
indicating better resistance to heat penetration.

\medskip
\noindent
\textbf{Dynamic thermal metrics.}
The periodic-day formulation additionally yields the two
standard dynamic characteristics of ISO~13786~\cite{ISO13786},
evaluated here in their simulation-based form with the sol-air
temperature \eqref{eq:solair} as the reference forcing signal.
The thermal time lag is the delay between the sol-air peak and
the inner-surface peak,
\begin{equation}
    \varphi(\boldsymbol{\mu})
    = \bigl[
        t_{\mathrm{peak}}\bigl(T(L,\cdot)\bigr)
        - t_{\mathrm{peak}}\bigl(T_{sa}\bigr)
      \bigr] \bmod \SI{24}{\hour},
    \label{eq:lag}
\end{equation}
and the decrement factor is the ratio of the inner-surface
swing to the sol-air swing over the final day,
\begin{equation}
    f(\boldsymbol{\mu})
    =
    \frac{\max_t T(L,t) - \min_t T(L,t)}
         {\max_t T_{sa}(t) - \min_t T_{sa}(t)}.
    \label{eq:decrement}
\end{equation}
A long time lag pushes the indoor heat peak into the cooler
night hours, and a small decrement factor indicates strong
attenuation of the external thermal wave; both complement the
peak temperature $J(\boldsymbol{\mu})$ in characterising
envelope performance under cyclic loading.
The five candidate indigenous materials and their nominal
properties are listed in Table~\ref{tab:materials}.

\begin{table}[htbp]
\centering
\caption{Candidate indigenous Sindh wall materials
         (nominal thermal properties)~\cite{Jannat2020,Oti2010,
         Laaroussi2014,ElAzhary2017,Shah2016,
         Koci2017,Adam1995,Cui2018}.}
\label{tab:materials}
\begin{tabular}{lcccc}
\toprule
Material
    & $k_{\mathrm{dry}}$ & $\rho$ & $c_p$ & $\alpha$ \\
    & (W\,m$^{-1}$K$^{-1}$) & (kg\,m$^{-3}$) & (J\,kg$^{-1}$K$^{-1}$) & (W\,m$^2$\,kg$^{-1}$K$^{-1}$) \\
\midrule
Mud brick (unfired)          & 0.45 & 1600 & 880  & 0.005 \\
Clay--straw adobe            & 0.35 & 1400 & 1000 & 0.004 \\
Lime-stabilised bamboo panel & 0.25 & 900  & 1800 & 0.003 \\
Fired clay brick             & 0.65 & 1800 & 840  & 0.005 \\
Lime--mud composite          & 0.55 & 1500 & 900  & 0.004 \\
\bottomrule
\end{tabular}
\end{table}

\noindent
The five candidate materials are selected on three grounds.
First, all five are constructible from raw materials native
to rural Sindh: earth, straw, lime, bamboo, and fired clay
are all locally available~\cite{Ebrahim2024,Iqbal2025}.
Fired clay brick in particular is produced at kilns operating
in Khairpur, Sukkur, and Larkana districts of Sindh.
The composite materials (clay--straw adobe, lime-stabilised
bamboo panel, lime--mud composite) are assembled from these
locally sourced inputs.
Mud brick (unfired) thermal properties are consistent with
values measured for unfired clay bricks, whose design
conductivities span approximately
$0.16$--$0.35$~W\,m$^{-1}$K$^{-1}$ over the studied density
range $\rho = 1200$--$1700$~kg\,m$^{-3}$, rising toward
$0.54$~W\,m$^{-1}$K$^{-1}$ at the upper end of the reported
density and moisture range~\cite{Oti2010},
and the adopted conductivity of clay--straw adobe
($k_\mathrm{dry} = 0.35$~W\,m$^{-1}$K$^{-1}$) is consistent
with the 48\% reduction relative to pure clay reported at
5\% straw mass fraction~\cite{ElAzhary2017}, with density
and specific heat assigned at values typical for earthen
composites within the ranges documented across adobe and
compressed earth blocks~\cite{Lachheb2023,Giada2019}.
The lime-stabilised bamboo panel represents an emerging
candidate, exemplified by post-2022 flood reconstruction
projects using bamboo-lime-mud construction in rural
Sindh~\cite{Ebrahim2024} and recommended for wider adoption
by Iqbal et al.~\cite{Iqbal2025};
it is included here as an emerging option rather than a
long-established traditional material.
The adopted dry thermal conductivity of
\SI{0.25}{\watt\per\metre\per\kelvin} for this panel is
bracketed by independent measurements of engineered bamboo
composites at comparable density: at the same density
(\SI{900}{\kilogram\per\cubic\metre}), Shah et
al.~\cite{Shah2016} report a parallel-grain conductivity of
approximately \SI{0.32}{\watt\per\metre\per\kelvin}, while
guarded hot-plate measurements at slightly lower densities
(682--856~\si{\kilogram\per\cubic\metre}) report conductivities
of 0.11--0.25~\si{\watt\per\metre\per\kelvin} across orthogonal
and parallel grain directions~\cite{Wang2018}; the isotropic
value adopted here falls within this orientation-dependent
range.
The density ($\rho = 900$~kg\,m$^{-3}$) is consistent with
the panel densities reported by Shah et al.~\cite{Shah2016}
for engineered bamboo composites produced at comparable
compaction levels.
The specific heat capacity ($c_p = 1800$~J\,kg$^{-1}$K$^{-1}$)
is consistent with differential scanning calorimetry (DSC) measurements of bamboo scrimber:
Cui et al.~\cite{Cui2018} report $c_p = 1570$~J\,kg$^{-1}$K$^{-1}$
at ambient temperature (20\,\textdegree C), rising to
1940~J\,kg$^{-1}$K$^{-1}$ at 320\,\textdegree C as moisture
is driven off and cellulose degrades; the adopted value of
\SI{1800}{\joule\per\kilogram\per\kelvin} falls within the
temperature-dependent range reported by Cui et
al.~\cite{Cui2018} and is representative of the material at
the elevated wall temperatures encountered during Sindh
summer loading.
No experimental measurement of $c_p$ for a lime-stabilised
bamboo panel assembly of the specific construction type
used in rural Sindh was identified in the literature,
and this residual uncertainty is bounded by the parametric
sweep over $c_p$ within the nine-dimensional LHS dataset.
The fired clay brick properties are corroborated by
independent experimental measurements of solid clay brick.
Koci et al.~\cite{Koci2017} report a dry-state density of
\SI{1831}{\kilogram\per\cubic\metre} and a dry-state thermal
conductivity of approximately
\SI{0.59}{\watt\per\metre\per\kelvin} for solid clay brick,
closely matching the density adopted here
(\SI{1800}{\kilogram\per\cubic\metre}).
Laaroussi et al.~\cite{Laaroussi2014} report a thermal
conductivity of approximately
\SI{0.35}{\watt\per\metre\per\kelvin} for fired clay brick.
The adopted conductivity of
\SI{0.65}{\watt\per\metre\per\kelvin} lies slightly above
the higher of these two independently measured values,
consistent with the documented sensitivity of fired-clay
conductivity to mineral phase composition and porosity
developed during sintering~\cite{Anjum2019}, and with the
firing temperatures of approximately 1000~\textdegree C
typical of the coal-fired kilns producing commercial brick
in Pakistan~\cite{Ahmad2008,Anjum2019}, well above the
780~\textdegree C firing of the bricks measured by
Laaroussi et al.~\cite{Laaroussi2014}; coal-baked commercial
brick sintered at approximately 1000~\textdegree C in
Pakistani kilns measures
$0.57 \pm 0.04$~W\,m$^{-1}$K$^{-1}$~\cite{Anjum2019},
comparable to the value adopted here.
The specific heat capacity ($c_p = 840$~J\,kg$^{-1}$K$^{-1}$)
is consistent with the range of 800--900~J\,kg$^{-1}$K$^{-1}$
reported for fired clay products in the building materials
literature~\cite{Jannat2020,Koci2018}.
Second, the five materials span the practical range of dry
thermal conductivity
($0.25$--$0.65$~\si{\watt\per\metre\per\kelvin})
and density ($900$--$1800$~\si{\kilogram\per\cubic\metre})
found in indigenous Sindh earthen and fired construction,
ensuring scientific breadth in the
comparison~\cite{Oti2010,Laaroussi2014,ElAzhary2017,Shah2016}.
Third, comparing multiple candidate materials within a single
framework is consistent with comparable parametric thermal
simulation studies in hot climates, such as the four-material
comparison of Jannat et al.~\cite{Jannat2020}.
The moisture--conductivity coupling coefficient $\alpha$ is
not tabulated for any of the five materials in the open
literature; nominal values are assigned by the authors at the
same order of magnitude as the coefficient reported by Wei
et al.~\cite{Wei2019} for a moisture-sensitive cementitious
wall material, and uncertainty is bounded by the parametric
sweep over $\alpha \in [0.001, 0.010]$ (Table~\ref{tab:params}).
The lime--mud composite, a locally practised wall assembly
in rural Sindh combining hydrated lime with Indus alluvial
clay, does not appear by name in the international thermal
properties literature.
The adopted values
($k_\mathrm{dry} = 0.55$~W\,m$^{-1}$K$^{-1}$,
$\rho = 1500$~kg\,m$^{-3}$, $c_p = 900$~J\,kg$^{-1}$K$^{-1}$)
are assigned by the authors on the basis of constituent
material bounds from the closest comparable materials
in the literature.
For thermal conductivity and density: lime-stabilised
and cement-stabilised hollow and plain earth blocks
produced from Sudanese soils report
$k = 0.25$--$0.55$~W\,m$^{-1}$K$^{-1}$ (oven-dry)
and specific heat capacity near
836~J\,kg$^{-1}$K$^{-1}$~\cite{Adam1995},
and lime-stabilised earthen blocks more broadly
span $k = 0.3$--$0.7$~W\,m$^{-1}$K$^{-1}$ and
$\rho = 1400$--$1800$~kg\,m$^{-3}$~\cite{Oti2010,Lachheb2023};
the adopted $k_\mathrm{dry} = 0.55$~W\,m$^{-1}$K$^{-1}$
and $\rho = 1500$~kg\,m$^{-3}$ fall within these ranges.
For specific heat: the adopted value of
900~J\,kg$^{-1}$K$^{-1}$ is consistent with the
836~J\,kg$^{-1}$K$^{-1}$ reported for stabilised earth
blocks~\cite{Adam1995} and the range of
800--900~J\,kg$^{-1}$K$^{-1}$ documented for
earthen and fired clay construction
materials~\cite{Jannat2020,Koci2018}, with the
small positive offset reflecting the higher lime fraction
in the Sindh composite relative to the stabilised blocks
in those studies.
The trained operator $\mathcal{G}_\theta$ can evaluate
additional materials not listed here by querying with their
thermophysical properties, without retraining.

\section{Methodology: FDM Data Generation and PINO Training}
\label{sec:method}

The computational framework proceeds in two stages: a validated
Crank--Nicolson FDM solver generates a dataset of high-fidelity
periodic-day solutions across the parameter space of
Section~\ref{sec:params}, and a physics-informed neural operator
is then trained on this dataset to serve as a fast surrogate.
This section motivates the two-stage design and details each
stage in turn.

\subsection{Rationale for the Two-Stage Approach}
\label{sec:why}

Evaluating wall thermal performance across hundreds of material
and climate combinations by running a full FDM simulation for
each combination is computationally expensive.
The two-stage approach resolves this: FDM generates a dataset
of high-fidelity solutions, and a Physics-Informed Neural
Operator (PINO) trained on this dataset evaluates any new
$\boldsymbol{\mu}$ in approximately 3~ms on a commodity GPU.
Unlike a purely data-driven FNO, PINO also enforces the
governing PDE \eqref{eq:heat} as a soft constraint in the
training loss, improving physical consistency
and reducing the number of FDM samples required.

\subsection{Stage 1: FDM Data Generation}
\label{sec:fdm}

The Finite Difference Method (FDM) solves the 1D heat
equation \eqref{eq:heat} on a uniform grid
$\{x_i\}_{i=0}^{N}$ with spacing $\Delta x = L / N$.
FDM is chosen because the 1D domain is simple, the parametric
sweep is easily scripted in Python, and full code ownership
enables transparent verification.

\medskip
\noindent
\textbf{Discretisation scheme.}
The heat equation is discretised using Crank--Nicolson
time-stepping~\cite{Morton1994} (second-order accurate,
unconditionally stable for diffusion equations) with
central differences in space; the Robin conditions
\eqref{eq:bc_out}--\eqref{eq:bc_in} are imposed to second
order via ghost-node elimination.
Because the material coefficients are constant within each run,
the time-varying forcing $T_{\mathrm{out}}(t)$ and
$Q_{\mathrm{solar}}(t)$ enters only the right-hand side of the
linear system; to preserve second-order temporal accuracy it is
averaged over the old and new time levels in the
Crank--Nicolson update.
The constant tridiagonal system matrix is lower--upper (LU) factorised once
per run and reused at every time step, reducing the cost of the
full 1500-sample sweep to minutes.
The production grid uses $N = 64$ spatial intervals (65 nodes),
4000 time steps per day ($\Delta t = \SI{21.6}{\second}$),
and stores the final day at 121 uniformly spaced instants.
At the start of each simulation run, $k_{\mathrm{eff}}$ is
computed once from \eqref{eq:keff} using the assigned values
of $k_{\mathrm{dry}}$, $\alpha$, and $w_0$; it remains
constant throughout the time loop.

\medskip
\noindent
\textbf{Dataset generation.}
Latin Hypercube Sampling (LHS) \cite{McKay1979} generates
$N_{\mathrm{s}} = 1500$ parameter vectors
$\{\boldsymbol{\mu}_i\}_{i=1}^{N_{\mathrm{s}}}$ from the
ranges in Table~\ref{tab:params}.
LHS ensures uniform coverage of the 9-dimensional parameter
space with far fewer samples than random Monte Carlo sampling.
For each $\boldsymbol{\mu}_i$, the FDM solver simulates five
forcing days and stores the final periodic day
$T_i(x, t)$ on a $65 \times 121$ space--time grid, together
with the per-sample dynamic metrics
\eqref{eq:lag}--\eqref{eq:decrement} and the periodicity
metric of Section~\ref{sec:ics}, forming the dataset:
\begin{equation}
    \mathcal{D}
    = \bigl\{
        \bigl(\boldsymbol{\mu}_i,\;
              T_i\bigr)
      \bigr\}_{i=1}^{N_{\mathrm{s}}}
    \label{eq:dataset}
\end{equation}
The dataset is split 80/10/10 into training
($N_{\mathrm{train}} = 1200$), validation
($N_{\mathrm{val}} = 150$), and held-out test
($N_{\mathrm{test}} = 150$) sets before any model training,
using a fixed random permutation seeded with
\texttt{seed=0}.

\subsection{Stage 2: PINO Model and Training}
\label{sec:pino}

In contrast to a conventional neural network, which
approximates a single function, a neural operator approximates
a \emph{mapping between function spaces}, in this case from the
parameter vector to the entire space--time temperature field,
so that a single trained model replaces the FDM solver
across the whole parameter range.
The Physics-Informed Neural Operator (PINO)~\cite{Li2021PINO} learns the
parameter-to-solution operator:
\begin{equation}
    \mathcal{G}_{\theta}
    \;:\;
    \boldsymbol{\mu}
    \;\longmapsto\;
    T(x,t)
    \label{eq:pino_op}
\end{equation}
where $\theta$ denotes the trainable network parameters,
using the FNO architecture \cite{Li2021FNO} as its backbone.
The parameter vector $\boldsymbol{\mu}\in\mathbb{R}^{9}$
(Table~\ref{tab:params}) is fed into the FNO by broadcasting
each of the nine scalar entries as a constant spatial channel
appended to the grid-point coordinate at every spatial node
$x_i$, following standard practice for parametric neural
operators, as implemented in the reference FNO
codebase~\cite{Li2021FNO}.
Specifically, the FNO input at grid point $(\xi_i,\tau_j)$ is the concatenated
vector $(\xi_i,\,\tau_j,\,\mu_1,\,\ldots,\,\mu_{9})\in\mathbb{R}^{11}$;
the spectral convolution layers then map this 11-channel input
over the $(x,t)$ grid to the output field
$T(x,t)$ at all saved time steps.
PINO is preferred over a purely data-driven FNO because it
additionally enforces the governing PDE as soft constraint
during training, which improves physical consistency and
reduces the required number of FDM training samples.
This physics-enforcement strategy is consistent with the
authors' prior work on physics-informed neural networks
for PDE-governed systems~\cite{Khan2026PINN}.

\medskip
\noindent
\textbf{Training loss.}
The total training loss combines a data loss and a PDE
residual loss:
\begin{equation}
    \mathcal{L}
    = \mathcal{L}_{\mathrm{data}}
    + \lambda_T\, \mathcal{L}_{T}^{\mathrm{PDE}}
    \label{eq:loss}
\end{equation}
The data loss matches PINO predictions to FDM solutions:
\begin{equation}
    \mathcal{L}_{\mathrm{data}}
    = \frac{1}{N_{\mathrm{train}}}
      \sum_{i=1}^{N_{\mathrm{train}}}
      \frac{\|\hat{T}_i - T_i\|_{L^2}}{\|T_i\|_{L^2}}
    \label{eq:loss_data}
\end{equation}
where $\hat{T}_i$ is the PINO prediction and $T_i$ is the FDM
solution for sample $i$, both evaluated on the full $(x,t)$ grid.
This relative $L^2$ form follows the \texttt{LpLoss} convention
used in the publicly released FNO and PINO implementations
of Li et al.~\cite{Li2021FNO,Li2021PINO}.
The PDE residual loss $\mathcal{L}_{T}^{\mathrm{PDE}}$
penalises violations of \eqref{eq:heat} at collocation
points inside the domain:
\begin{equation}
    \mathcal{L}_{T}^{\mathrm{PDE}}
    = \frac{1}{N_c}
      \sum_{j=1}^{N_c}
      \left(
        \rho\,c_p\,
        \frac{\partial \hat{T}}{\partial t}
        -
        \frac{\partial}{\partial x}
        \!\left(
          k_{\mathrm{eff}}\,
          \frac{\partial \hat{T}}{\partial x}
        \right)
      \right)^{\!2}_{\!(x_j,\,t_j)}
    \label{eq:loss_pde}
\end{equation}
where the sum is taken over all interior grid points
$(x_j, t_j) \in \{x_1,\ldots,x_{N-1}\} \times \{t_1,\ldots,t_{N_t}\}$,
excluding the boundary nodes $x=0$, $x=L$ (governed by Robin
conditions) and the initial time $t=0$ (governed by the initial condition).
The temporal and spatial derivatives of $\hat{T}$ are computed
by second-order central finite differences on the uniform
$(\xi,\tau)$ grid; this extends to both dimensions the
finite-difference-in-time strategy of Li et al.~\cite{Li2021PINO},
who compute spatial derivatives spectrally.
This avoids differentiating through the fast Fourier transform (FFT) layers of the
FNO backbone and is exact to $\mathcal{O}(\Delta\xi^2,
\Delta\tau^2)$ on the uniform grid.
The weight $\lambda_T$ balances the data and physics terms
and is determined by tuning.
Because the stored window is the final periodic day, it
contains no start-up transient; the finite-difference residual
of the exact FDM fields on the $65 \times 121$ grid is
$\sim 3 \times 10^{-3}$ in the non-dimensionalised form,
setting a small irreducible floor for
$\mathcal{L}_{T}^{\mathrm{PDE}}$.

\medskip
\noindent
\textbf{Training setup.}
The FNO backbone uses 4 spectral convolution layers with
channel width 32, Fourier mode cutoffs $M_x = 16$ and
$M_t = 20$ in the spatial and temporal dimensions
respectively, and Gaussian Error Linear Unit (GELU) activations.
The model is trained in PyTorch (Google Colab, T4 graphics processing unit, GPU) using
the Adam optimiser with initial learning rate $10^{-3}$ and
weight decay $10^{-5}$,
reduced by a factor of 0.5 when the validation loss has not
improved for 10 consecutive epochs (minimum lr $10^{-6}$),
with early stopping after 30 epochs of no improvement and
a maximum of 300 epochs.
Mini-batches of size 64 are used, with random shuffling
of the training set each epoch; the \texttt{drop\_last=True} flag
ensures uniform batch sizes throughout training, and gradient
norms are clipped at 1.0.
The PDE weight is set to $\lambda_T = 0.01$, selected by a
pilot grid search over $\lambda_T \in \{0.001, 0.01, 0.1\}$.
Immediately before initialisation, \texttt{torch.manual\_seed(42)}
is set, so that both the FNO baseline and the PINO model start
from identical weight initialisations and an identical
training-set shuffling order, isolating the loss function as
the only difference between the two training runs.
The acceptance criterion is a relative $L^2$ error below 5\%
on the temperature field $T$ on the held-out test set.
Relative $L^2$ error on $T$ and mean absolute error (MAE) on the
QoI $J$ are reported in Table~\ref{tab:pino_acc}.

\section{Results}
\label{sec:results}

The FDM solver achieves second-order convergence in both the
manufactured-solution test and the time-varying diurnal
problem (\SI{0.58}{\milli\kelvin} inner-surface error at
production resolution), passes the Robin BC zero-drift test to
within $10^{-12}$\,K, and reaches a verified periodic
quasi-steady state on the extracted final day
(Section~\ref{sec:fdm_val}).
The trained PINO attains a relative $L^2$ field error of
$5.14\times10^{-4}$ and a \SI{0.201}{\kelvin} MAE on the peak
inner surface temperature $J(\boldsymbol{\mu})$, reproducing
the FDM material ranking exactly; the physics loss is most
valuable near the inner surface, where it reduces the root-mean-square error (RMSE) by
15.5\%, and in the data-scarce regime, where PINO trained on
150 FDM samples matches a data-only FNO trained on 300
(Sections~\ref{sec:pino_acc}--\ref{sec:data_eff}).

\subsection{FDM Solver Validation}
\label{sec:fdm_val}

Before generating the full dataset, the FDM solver is
validated against the analytical steady-state solution for a
uniform wall with constant $k$, zero moisture ($w = 0$), and
Dirichlet boundary conditions $T(0) = T_{\mathrm{out}}$,
$T(L) = T_{\mathrm{in}}$.
The exact solution is linear in $x$; the FDM error is
expected to be $\mathcal{O}(\Delta x^2)$.
Because a transient run does not reach this steady state
within the integration window for the materials and wall
thicknesses considered (Section~\ref{sec:ics}), the spatial
convergence test instead uses the method of manufactured
solutions~\cite{Roache2002,Salari2000}:
the exact transient field
$T(x,t) = T^*(x) + \sin(\pi x/L)\exp(-\alpha_{\mathrm{th}}(\pi/L)^2 t)$,
where $\alpha_{\mathrm{th}} = k/(\rho c_p)$ and $T_D^*(x) = T_{\mathrm{out}} + (T_{\mathrm{in}} - T_{\mathrm{out}})x/L$ is the Dirichlet steady state (distinct from the Robin steady-state profile $T^*(x)$ in \eqref{eq:T_robin}), satisfies the heat equation
exactly with the Dirichlet boundary values, and is used as
both the initial condition and the reference solution at
$t = 12$~h.

\medskip\noindent\textbf{Robin boundary condition validation.}
The Robin BC discretisation at $x = 0$, which carries the
solar heat flux $Q_{\mathrm{solar}}$ that drives all five
wall comparisons, is validated separately from the Dirichlet
test above, using the exact constant-forcing steady state
$T^*(x)$ \eqref{eq:T_robin} derived in
Section~\ref{sec:bc_heat}.
The FDM Robin BC implementation is confirmed via a
zero-drift test: with the forcing held constant, starting the
solver from the exact $T^*(x)$ \eqref{eq:T_robin} and
advancing one full day (24~h) leaves the field unchanged to
within $7.4\times10^{-13}$~K at all grid resolutions
$N \in \{16,32,64,128\}$, confirming that the discrete
Robin steady state matches $T^*(x)$ to machine precision.

Table~\ref{tab:fdm_conv} confirms second-order spatial convergence
($\mathcal{O}(\Delta x^2)$) at all four grid resolutions.

\begin{table}[htbp]
\centering
\caption{FDM grid-convergence study: Dirichlet validation via
         method of manufactured solutions (Section~\ref{sec:fdm_val}).}
\label{tab:fdm_conv}
\begin{tabular}{cccc}
\toprule
$N$ & $\Delta x$ (m) & $L^\infty$ error (K) & Order \\
\midrule
16  & 0.01562 & $9.97\times10^{-4}$ & -- \\
32  & 0.00781 & $2.49\times10^{-4}$ & 2.00 \\
64  & 0.00391 & $6.22\times10^{-5}$ & 2.00 \\
128 & 0.00195 & $1.55\times10^{-5}$ & 2.00 \\
\bottomrule
\end{tabular}
\end{table}

\medskip\noindent\textbf{Time-varying transient convergence.}
The manufactured-solution and zero-drift tests exercise fixed or
steady conditions; the diurnal problem that generates the
dataset is additionally verified by simultaneous space--time
refinement against a fine reference solution
($N = 256$, 32{,}000 steps per day) at the nominal parameter
vector.
Table~\ref{tab:fdm_conv_diurnal} reports the maximum
inner-surface error over the final day and the peak-temperature
error at four resolutions refined jointly in $\Delta x$ and
$\Delta t$: the observed convergence orders are 2.01--2.32, and
the production settings ($N = 64$, 4000 steps per day) resolve
the final-day inner-surface trace to
\SI{0.58}{\milli\kelvin}, three orders of magnitude below
the PINO error budget of Section~\ref{sec:pino_acc}.

\begin{table}[htbp]
\centering
\caption{FDM space--time convergence for the diurnal
         time-varying problem at the nominal parameter vector,
         against a fine reference ($N=256$, 32{,}000 steps/day).
         Errors are evaluated on the final-day inner-surface
         trace $T(L,t)$.}
\label{tab:fdm_conv_diurnal}
\begin{tabular}{cccc}
\toprule
$N$ & Steps/day & $\max_t |T(L,t) - T_{\mathrm{ref}}(L,t)|$ (K)
    & $|J_{\mathrm{peak}} - J_{\mathrm{peak,ref}}|$ (K) \\
\midrule
16  & 1000 & $9.88\times10^{-3}$ & $3.66\times10^{-3}$ \\
32  & 2000 & $2.45\times10^{-3}$ & $9.16\times10^{-4}$ \\
64  & 4000 & $5.83\times10^{-4}$ & $2.19\times10^{-4}$ \\
128 & 8000 & $1.17\times10^{-4}$ & $4.38\times10^{-5}$ \\
\bottomrule
\end{tabular}
\end{table}

\medskip\noindent\textbf{Periodicity and initial-condition
independence.}
The claim that the extracted final day represents the periodic
quasi-steady state (Section~\ref{sec:ics}) is verified on the
most sluggish wall in the parameter space (lime-stabilised
bamboo panel at $L = \SI{0.45}{\metre}$): with the mean-forcing
initial condition \eqref{eq:ic_T}, the day-4 to day-5
inner-surface difference is
$1.25\times10^{-2}$\,K, whereas a uniform \SI{305}{\kelvin}
initial condition leaves a residual of
$1.61\times10^{-1}$\,K after the same five-day spin-up,
demonstrating both that the mean-forcing initialisation
accelerates convergence to the periodic attractor and that the
reported final day is insensitive to the initial condition once
converged.
Across the full 1500-sample production sweep the periodicity
metric (day-to-day inner-surface temperature difference) does not exceed $4.3\times10^{-2}$\,K (median
$2.0\times10^{-3}$\,K), well below the surrogate error budget.

\subsection{PINO Accuracy}
\label{sec:pino_acc}
\begin{figure}[htbp]
\centering
\includegraphics[width=0.7\textwidth]{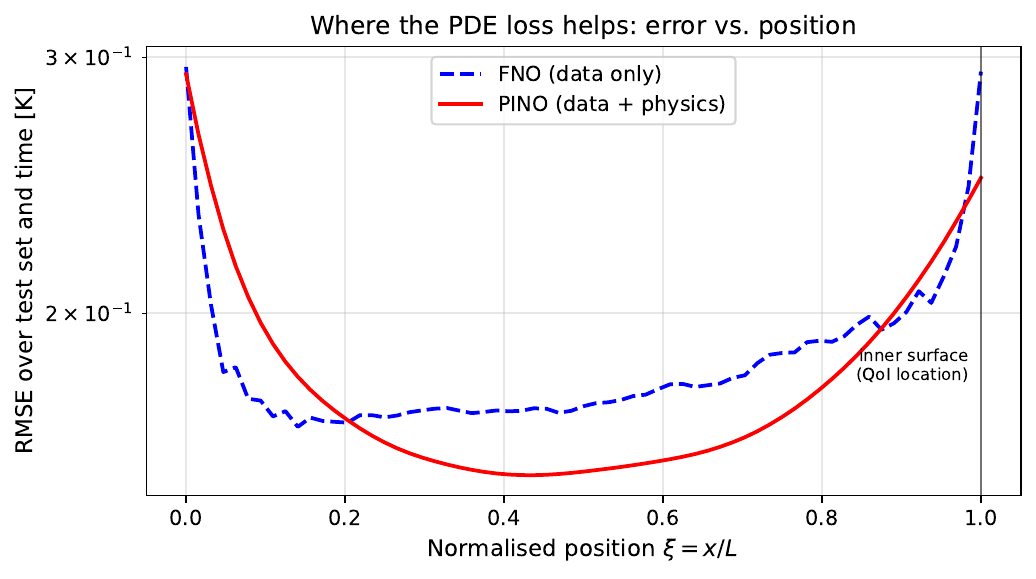}
\caption{Test-set RMSE (over samples and time) as a function of
         normalised position $\xi = x/L$ for the FNO baseline
         and PINO. The physics loss preferentially reduces the
         error near the inner surface $\xi = 1$, the location
         where the QoI \eqref{eq:qoi_peak} is evaluated
         (15.5\% RMSE reduction).}
\label{fig:err_position}
\end{figure}

The PINO model improves on the data-only FNO baseline across both accuracy metrics.
Table~\ref{tab:pino_acc} compares PINO against a purely
data-driven FNO baseline trained on the same dataset under the
protocol of Section~\ref{sec:pino}: identical weight
initialisation and data ordering, with validation-loss early
stopping selecting epoch~176 for the FNO and epoch~205 for the
PINO.
On the full 1200-sample training set, PINO achieves a relative
$L^2$ error of $5.14\times10^{-4}$ on the temperature field
$T$, a $5.1\%$ improvement over the FNO baseline
($5.42\times10^{-4}$), and reduces MAE on the scalar QoI
$J(\boldsymbol{\mu})$~\eqref{eq:qoi_peak} from
$0.205$\,K to $0.201$\,K.
Both models satisfy the $5\%$ relative $L^2$ acceptance
criterion (Section~\ref{sec:method}) by more than two orders
of magnitude.
At this data volume the two models are thus comparable in
aggregate accuracy, as expected for a smooth, linear
one-dimensional operator that an FNO can already learn well
from 1200 samples, and the value of the physics loss appears
instead in \emph{where} the error is reduced and in \emph{how
much data} is required.
Figure~\ref{fig:err_position} resolves the test-set
root-mean-square error as a function of position,
$\mathrm{RMSE}(\xi) = \bigl[\tfrac{1}{N_{\mathrm{test}}\,N_t}
\sum_{n,j}\bigl(\hat{T}_n(\xi,\tau_j)-T_n(\xi,\tau_j)\bigr)^2
\bigr]^{1/2}$ in kelvin: the PDE residual loss preferentially
improves accuracy near the inner wall surface, reducing the
RMSE at $\xi = 1$, the location where the QoI is evaluated,
from \SI{0.293}{\kelvin} (FNO) to \SI{0.248}{\kelvin}
(PINO), a $15.5\%$ reduction.
The data-efficiency question is addressed in
Section~\ref{sec:data_eff}.

\begin{table}[htbp]
\centering
\caption{PINO vs.\ FNO baseline accuracy on
         $\mathcal{D}_{\mathrm{test}}$ ($N_{\mathrm{test}}=150$,
         split via \texttt{seed=0}).
         Both models initialised with \texttt{torch.manual\_seed(42)}
         immediately before training, ensuring identical weight
         initialisation and shuffling order;
         FNO: data-only loss, early stopped at epoch~176;
         PINO: $\lambda_T=0.01$, early stopped at epoch~205.}
\label{tab:pino_acc}
\begin{tabular}{lcc}
\toprule
Model & Rel.\ $L^2$ on $T$ & MAE on $J$ (K) \\
\midrule
FNO (data only) & $5.42\times10^{-4}$ & $0.205$ \\
PINO (ours)     & $5.14\times10^{-4}$ & $0.201$ \\
\bottomrule
\end{tabular}
\end{table}

\begin{figure}[htbp]
\centering
\includegraphics[width=\textwidth]{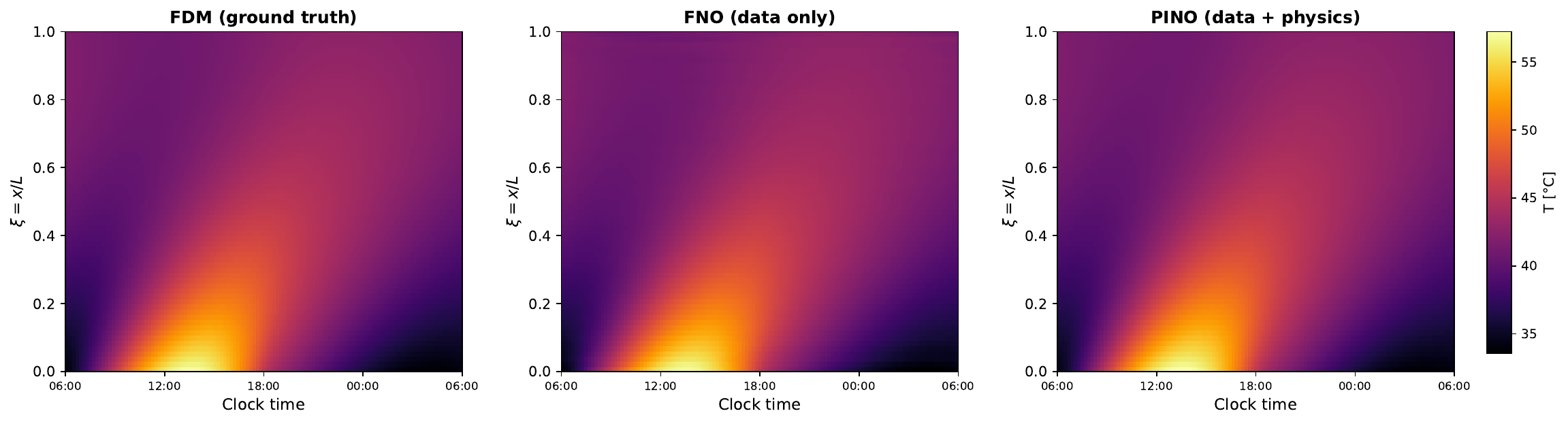}
\caption{Temperature field $T(\xi,t)$ over the final periodic
         day for the median-PINO-error test sample
         ($\mathcal{D}_{\mathrm{test}}$, sample index 28):
         FDM ground truth (left), FNO prediction (middle), and
         PINO prediction (right). All three panels share a common
         colour scale and no clipping is applied to the
         predictions; the time axis shows clock time from
         sunrise (06:00) to sunrise.}
\label{fig:field_comparison}
\end{figure}

\begin{figure}[htbp]
\centering
\includegraphics[width=0.7\textwidth]{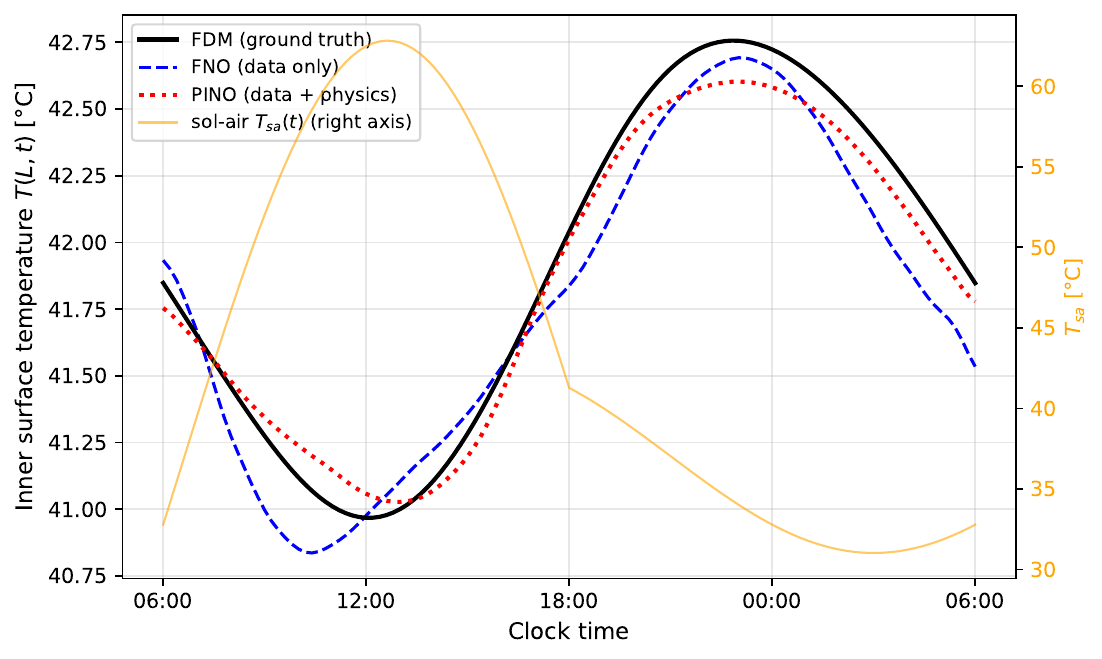}
\caption{Inner surface temperature $T(L,t)$ over the final
         periodic day, the time series $J(t;\boldsymbol{\mu})$
         \eqref{eq:qoi} underlying the scalar QoI
         $J(\boldsymbol{\mu})$ \eqref{eq:qoi_peak}, for
         the same median-error test sample as
         Figure~\ref{fig:field_comparison}, with the sol-air
         forcing $T_{sa}(t)$ \eqref{eq:solair} shown on the
         right-hand axis for context. The attenuation and phase
         shift of the inner-surface response relative to the
         forcing correspond to the decrement factor and time lag
         of \eqref{eq:lag}--\eqref{eq:decrement}.}
\label{fig:qoi_timeseries}
\end{figure}

\begin{figure}[htbp]
\centering
\includegraphics[width=\textwidth]{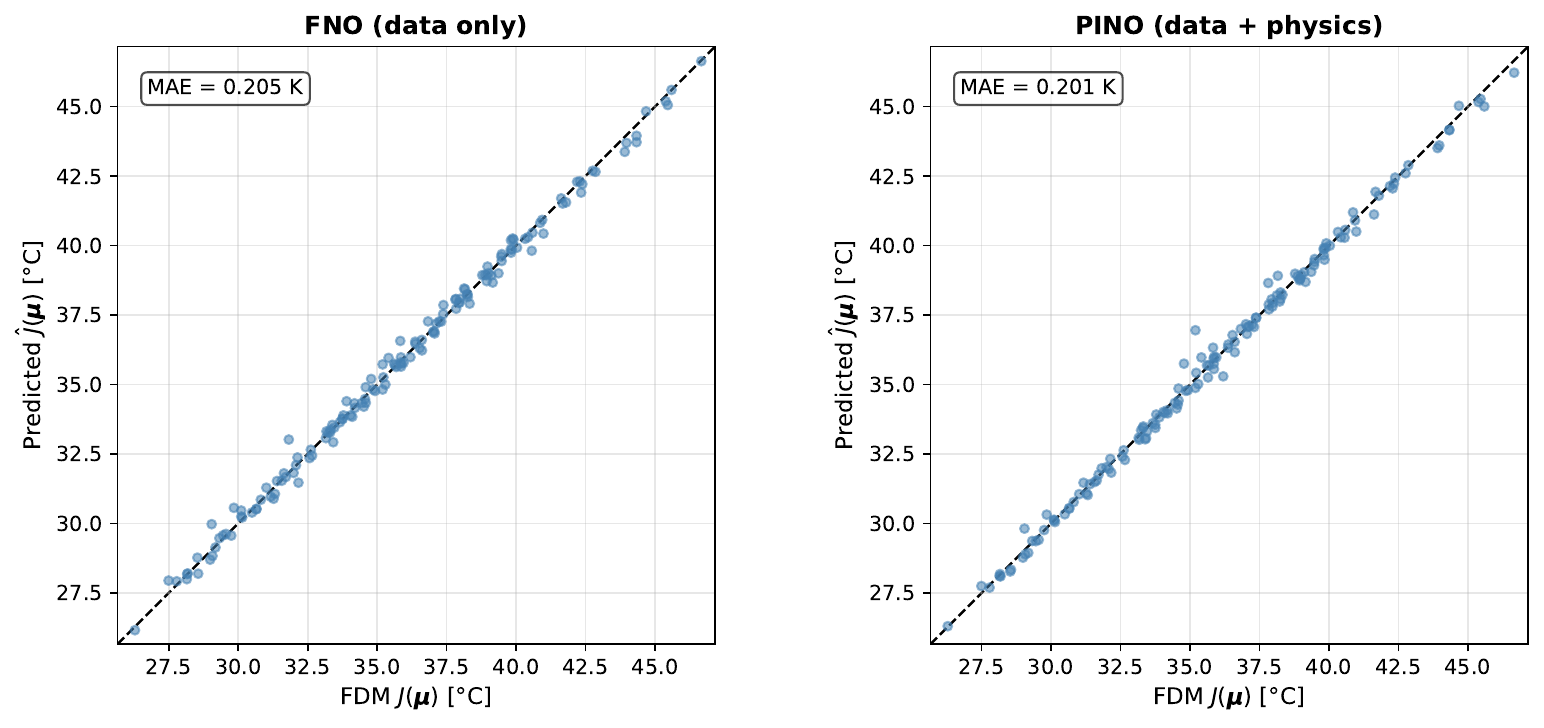}
\caption{Parity plots of the predicted peak inner surface
         temperature $\hat{J}(\boldsymbol{\mu})$ against the FDM
         ground truth $J(\boldsymbol{\mu})$ across the full
         held-out test set ($N_{\mathrm{test}}=150$), for FNO
         (left) and PINO (right). Both panels share identical axis
         limits for direct visual comparison; the dashed line
         denotes perfect agreement. The two panels are visually
         similar, consistent with PINO's modest ($\sim$2\%)
         relative improvement in MAE on $J$ over FNO
         (Table~\ref{tab:pino_acc}).}
\label{fig:parity}
\end{figure}

\begin{figure}[htbp]
\centering
\includegraphics[width=\textwidth]{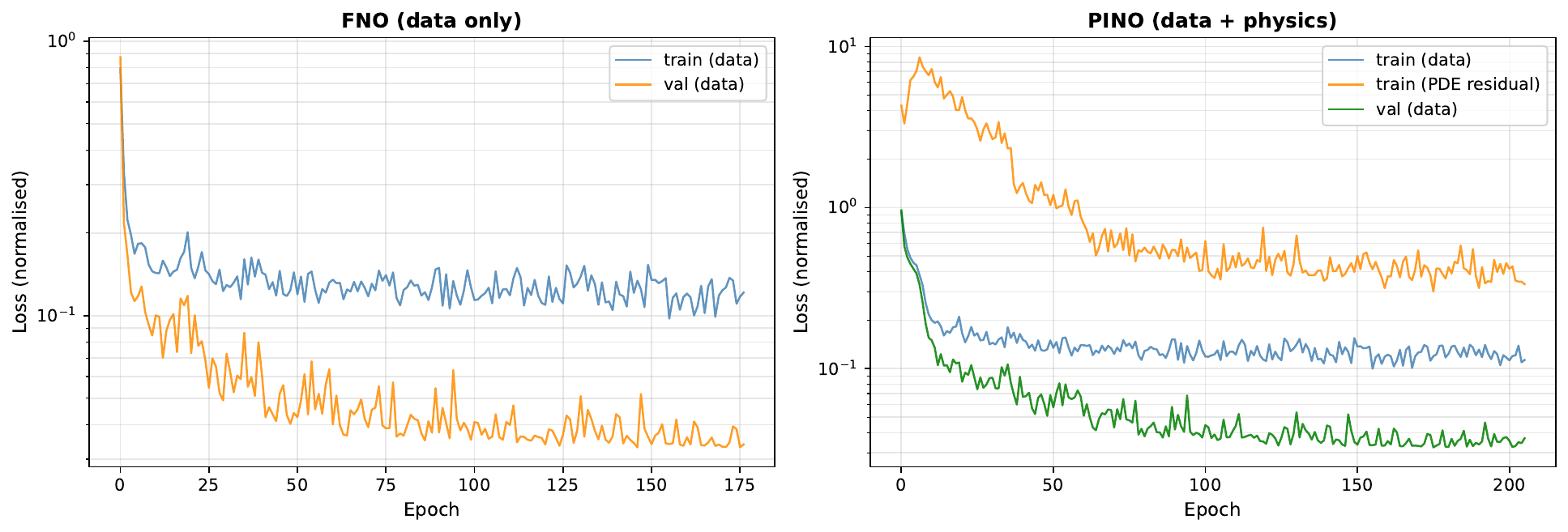}
\caption{Training and validation loss curves for the data-only
         FNO baseline (left) and the PINO model with
         $\lambda_T=0.01$ (right). Both losses are the relative
         $L^2$ error \eqref{eq:loss_data}; for PINO, the unweighted
         PDE residual term \eqref{eq:loss_pde} is shown
         separately. Both models were trained with the Adam
         optimiser (initial learning rate $10^{-3}$, with
         \texttt{ReduceLROnPlateau} (factor 0.5, patience 10),
         batch size 64), each re-initialised with
         \texttt{torch.manual\_seed(42)} immediately before
         training for a fair comparison, with early stopping
         after 30 epochs of no improvement on the validation
         loss, stopping at epoch~176 (FNO) and epoch~205 (PINO),
         as noted in Table~\ref{tab:pino_acc}.}
\label{fig:training_curves}
\end{figure}

\begin{figure}[htbp]
\centering
\includegraphics[width=0.7\textwidth]{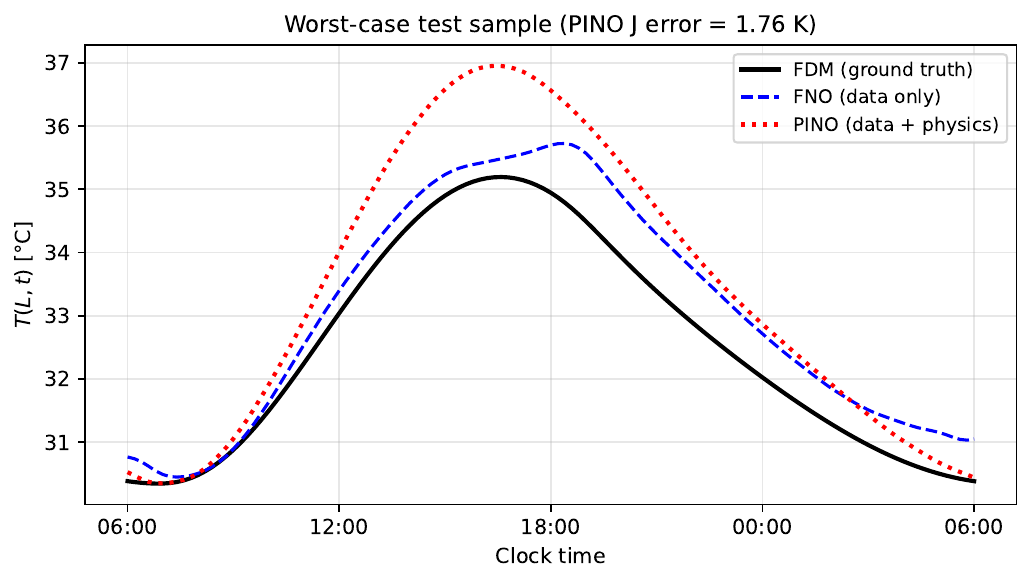}
\caption{Worst-case test sample for PINO (index 128, QoI error
         \SI{1.76}{\kelvin}): inner-surface temperature over
         the final periodic day for FDM, FNO, and PINO. Shown
         for transparency alongside the median-error sample of
         Figure~\ref{fig:qoi_timeseries}.}
\label{fig:worst_case}
\end{figure}

Figures~\ref{fig:field_comparison}, \ref{fig:qoi_timeseries},
\ref{fig:parity}, \ref{fig:training_curves},
and~\ref{fig:err_position} illustrate these results in detail.
The sample shown in
Figures~\ref{fig:field_comparison}--\ref{fig:qoi_timeseries}
is the \emph{median}-error test sample for PINO (index 28,
QoI error \SI{0.153}{\kelvin}), selected as the median of the
per-sample QoI error distribution to avoid favourable
selection; the worst-case test sample (index 128, QoI error
\SI{1.76}{\kelvin}) is shown in
Figure~\ref{fig:worst_case} for transparency.

\subsection{Data Efficiency}
\label{sec:data_eff}

\begin{figure}[htbp]
\centering
\includegraphics[width=\textwidth]{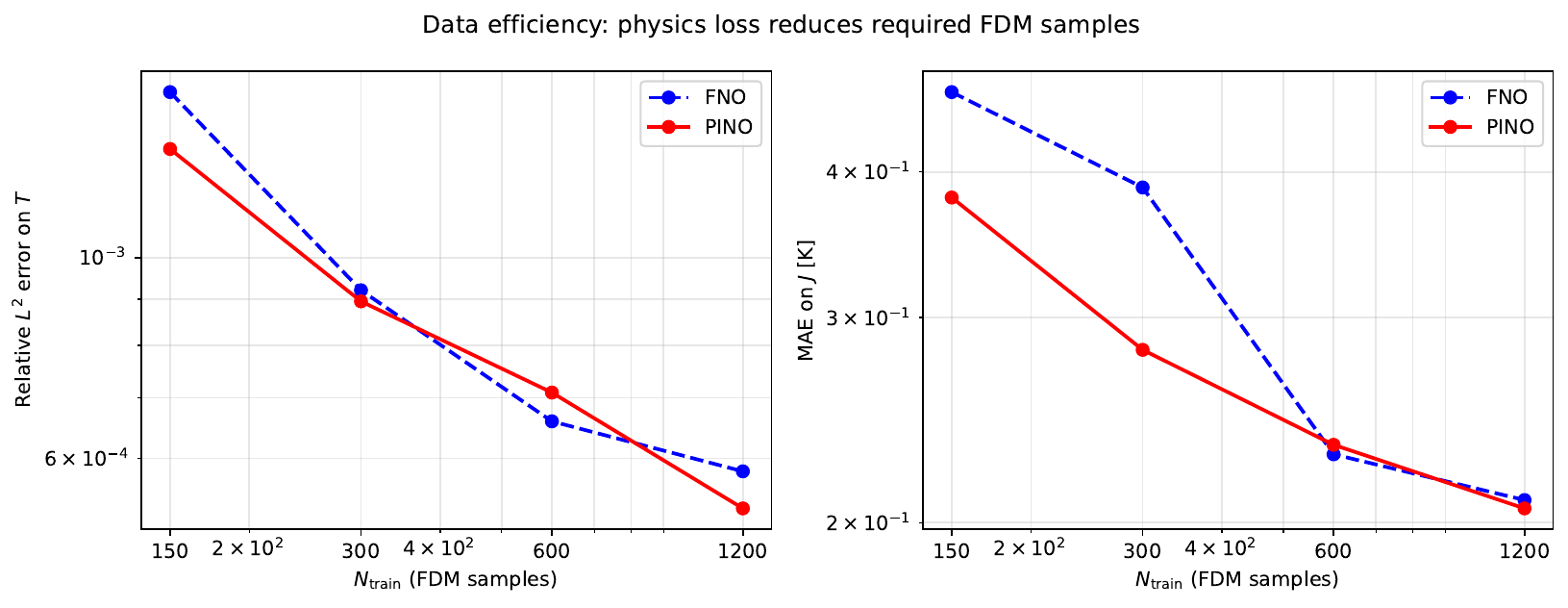}
\caption{Data efficiency of the physics loss: test-set relative
         $L^2$ error on $T$ (left) and MAE on $J$ (right) as a
         function of training-set size, for the data-only FNO
         and PINO under identical training conditions.}
\label{fig:data_eff}
\end{figure}

The claim that PDE enforcement reduces the number of
high-fidelity FDM samples required (Section~\ref{sec:pino})
is tested directly by retraining both models on nested subsets
of the training split ($N_{\mathrm{train}} \in
\{150, 300, 600, 1200\}$), with identical seeds,
normalisation statistics, validation and test sets, and a
200-epoch cap; results are reported on the fixed 150-sample
test set (Table~\ref{tab:data_eff},
Figure~\ref{fig:data_eff}).
The small difference from Table~\ref{tab:pino_acc} at
$N_{\mathrm{train}} = 1200$ reflects the shorter 200-epoch cap
used here.

\begin{table}[htbp]
\centering
\caption{Data-efficiency study: test-set accuracy vs.\
         training-set size for the data-only FNO and PINO
         ($\lambda_T = 0.01$), trained on nested subsets under
         identical conditions (200-epoch cap). The main-text
         models of Table~\ref{tab:pino_acc} use a 300-epoch cap,
         hence the small difference at
         $N_{\mathrm{train}} = 1200$.}
\label{tab:data_eff}
\begin{tabular}{ccccc}
\toprule
$N_{\mathrm{train}}$
    & \multicolumn{2}{c}{Rel.\ $L^2$ on $T$}
    & \multicolumn{2}{c}{MAE on $J$ (K)} \\
\cmidrule(lr){2-3}\cmidrule(lr){4-5}
    & FNO & PINO & FNO & PINO \\
\midrule
150  & $1.53\times10^{-3}$ & $1.32\times10^{-3}$ & 0.468 & \textbf{0.380} \\
300  & $9.21\times10^{-4}$ & $8.95\times10^{-4}$ & 0.388 & \textbf{0.281} \\
600  & $6.59\times10^{-4}$ & $7.09\times10^{-4}$ & \textbf{0.229} & 0.233 \\
1200 & $5.80\times10^{-4}$ & $5.28\times10^{-4}$ & 0.209 & \textbf{0.206} \\
\bottomrule
\end{tabular}
\end{table}

The physics loss is most valuable in the data-scarce regime:
at $N_{\mathrm{train}} = 150$ PINO reduces the QoI MAE by
$18.8\%$ (0.380\,K vs.\ 0.468\,K), and at
$N_{\mathrm{train}} = 300$ by $27.5\%$ (0.281\,K vs.\
0.388\,K); PINO trained on 150 samples matches the
accuracy of the data-only FNO trained on 300, so the physics
loss halves the FDM data budget at this accuracy level.
As the training set grows the advantage diminishes and the two
models converge. The single-run values at
$N_{\mathrm{train}} = 600$ (Table~\ref{tab:data_eff}) place FNO
marginally ahead, but this reflects seed sensitivity rather than
a genuine crossover: repeating both trainings over three seeds
$\{0, 42, 123\}$ gives an MAE on $J$ of
$0.246 \pm 0.026$~K for FNO and $0.234 \pm 0.008$~K for PINO,
so the two are statistically indistinguishable at this training
size (the difference is well within the combined seed spread).
In particular, the physics-constrained model is markedly less
sensitive to initialisation (its across-seed standard
deviation is roughly a third of the baseline's), indicating
that the PDE residual also stabilises training, a practical
benefit distinct from accuracy. This trajectory is consistent
with the interpretation of the PDE residual as a physics-based
regulariser whose information content is progressively
superseded by data.
The comparison is, if anything, conservative towards PINO: at
$N_{\mathrm{train}} \in \{150, 300\}$ the PINO runs reached
the 200-epoch cap with validation loss still improving, whereas
the corresponding FNO runs stopped early.

\subsection{Dynamic Thermal Metrics}
\label{sec:dyn_metrics}

\begin{figure}[htbp]
\centering
\includegraphics[width=\textwidth]{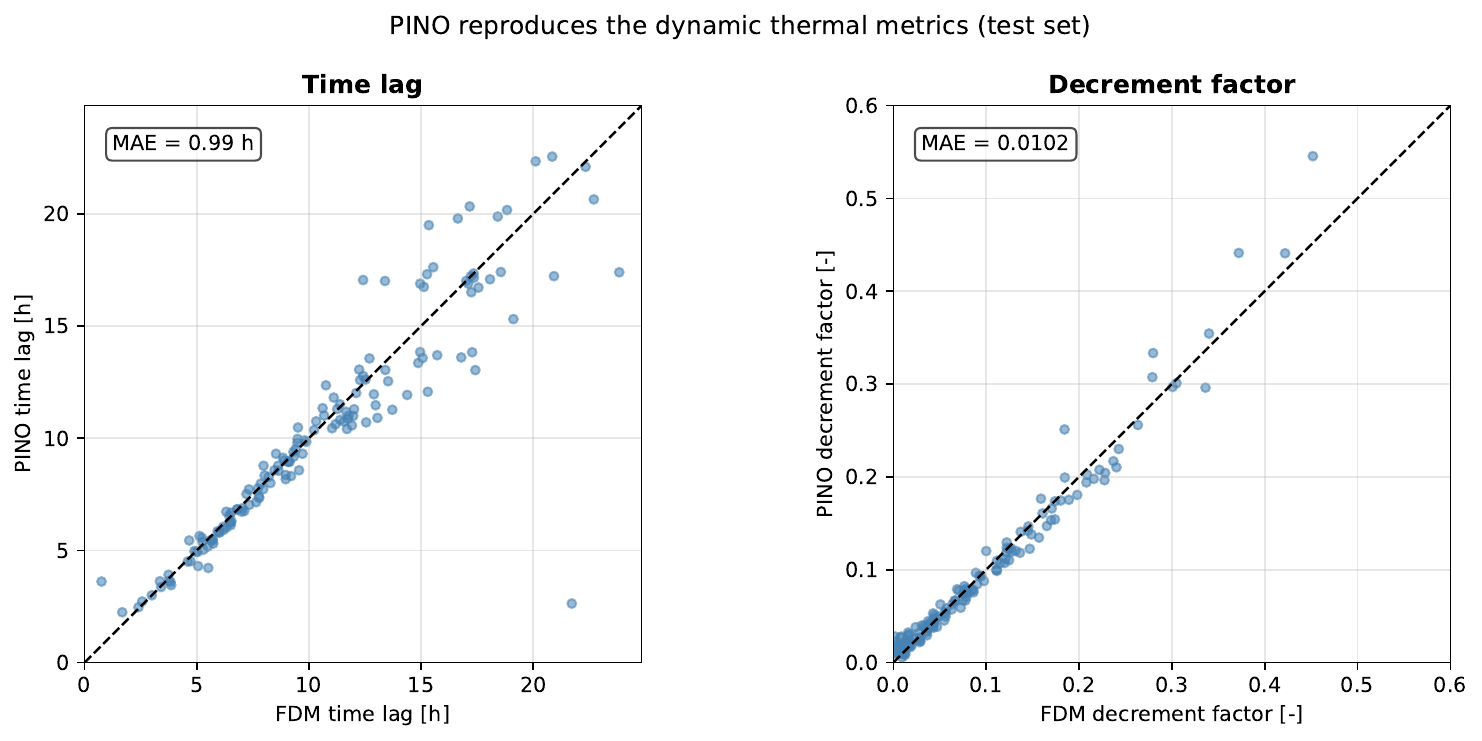}
\caption{Parity of the dynamic thermal metrics across the test
         set: time lag \eqref{eq:lag} (left) and decrement
         factor \eqref{eq:decrement} (right), PINO prediction
         vs.\ FDM ground truth, both evaluated on the same
         saved final-day grid.}
\label{fig:dyn_parity}
\end{figure}

Because the operator predicts the full space--time field, the
dynamic metrics \eqref{eq:lag}--\eqref{eq:decrement} can be
evaluated directly from PINO predictions.
Figure~\ref{fig:dyn_parity} compares PINO-derived time lags
and decrement factors against the FDM values across the full
test set, with both evaluated on the same 121-point saved grid
(three-point parabolic refinement of the peak location).
The operator reproduces the time lag with an MAE of
\SI{0.99}{\hour} (comparable to the 12-minute resolution of
the saved grid) and the decrement factor with an MAE of
0.010, confirming that the surrogate captures not only the peak
temperature but the full attenuation-and-delay character of the
envelope response.

\subsection{Material Ranking at Nominal Conditions}
\label{sec:ranking}
At nominal Sindh climate conditions, the five candidate materials produce distinct peak inner-surface temperatures.
Using the trained PINO $\mathcal{G}_\theta$,
$J(\boldsymbol{\mu})$ \eqref{eq:qoi_peak} is evaluated for all
five candidate materials (Table~\ref{tab:materials}) at nominal
Sindh climate conditions
($T_{\mathrm{out,max}} = \SI{45}{\celsius}$,
$T_{\mathrm{in}} = \SI{35}{\celsius}$,
$G_{s,\mathrm{peak}} = \SI{700}{\watt\per\metre\squared}$,
$w_0 = 0$~\si{\kilogram\per\cubic\metre},
$L = \SI{0.25}{\metre}$), on the final periodic day.
High-fidelity FDM solutions at the same conditions serve as
ground truth; PINO predictions are compared against them to
quantify prediction error on the ranking task, and the FDM
values are the basis of all downstream analyses.
Table~\ref{tab:ranking} reports both, together with the FDM
dynamic metrics \eqref{eq:lag}--\eqref{eq:decrement}.

\begin{table}[htbp]
\centering
\caption{Material ranking by peak inner surface temperature
         $J(\boldsymbol{\mu})$ \eqref{eq:qoi_peak} on the final
         periodic day at nominal Sindh conditions
         ($T_{\mathrm{out,max}}=\SI{45}{\celsius}$,
          $T_{\mathrm{in}}=\SI{35}{\celsius}$,
          $G_{s,\mathrm{peak}}=\SI{700}{\watt\per\metre\squared}$,
          $w_0=0$, $L=\SI{0.25}{\metre}$).
         FDM: direct solver output (ground truth).
         PINO: operator prediction $\mathcal{G}_\theta(\boldsymbol{\mu})$.
         Error: $|J_{\mathrm{PINO}} - J_{\mathrm{FDM}}|$.
         $\varphi$: thermal time lag \eqref{eq:lag} and
         $f$: decrement factor \eqref{eq:decrement}, both from
         FDM.
         Avail.: local availability in rural Sindh
         (\textbf{H}~=~widely available,
          \textbf{M}~=~regional,
          \textbf{L}~=~limited/emerging).}
\label{tab:ranking}
\begin{tabular}{clcccccc}
\toprule
Rank
    & Material
    & $J_{\mathrm{FDM}}$ (\si{\celsius})
    & $J_{\mathrm{PINO}}$ (\si{\celsius})
    & Error (K)
    & $\varphi$ (h)
    & $f$ (--)
    & Avail.\ \\
\midrule
1 & Lime-stabilised bamboo panel & 36.37 & 37.08 & 0.708 & 13.4 & 0.016 & \textbf{L} \\
2 & Clay--straw adobe            & 37.07 & 37.41 & 0.340 & 10.1 & 0.039 & \textbf{H} \\
3 & Mud brick (unfired)          & 37.69 & 37.92 & 0.228 & 8.8  & 0.058 & \textbf{H} \\
4 & Lime--mud composite          & 38.36 & 38.39 & 0.031 & 7.7  & 0.082 & \textbf{M} \\
5 & Fired clay brick             & 38.76 & 38.70 & 0.055 & 7.6  & 0.090 & \textbf{H} \\
\bottomrule
\end{tabular}
\end{table}

Lime-stabilised bamboo panel achieves the lowest peak inner
surface temperature ($J_{\mathrm{FDM}} = \SI{36.37}{\celsius}$),
together with the longest time lag ($\varphi = \SI{13.4}{\hour}$)
and the strongest attenuation ($f = 0.016$), owing to its low
thermal conductivity
($k_{\mathrm{dry}} = \SI{0.25}{\watt\per\metre\per\kelvin}$)
and high specific heat capacity
($c_p = \SI{1800}{\joule\per\kilogram\per\kelvin}$).
However, its limited local availability (\textbf{L}) constrains
its practical deployment in remote rural Sindh communities.
Among materials with wide local availability (\textbf{H}),
clay--straw adobe ranks highest thermally
($J_{\mathrm{FDM}} = \SI{37.07}{\celsius}$), only
$\SI{0.70}{\kelvin}$ warmer than the bamboo panel, with a
\SI{10.1}{\hour} time lag that displaces the indoor heat peak
into the late-evening hours, and at substantially lower cost
(Table~\ref{tab:cost}).
The dynamic metrics order identically to $J$: peak temperature,
time lag, and decrement factor all favour the same
low-diffusivity materials under cyclic loading.
The cost--performance assessment in Section~\ref{sec:cost}
incorporates these factors into a single index.
The maximum prediction error on the ranking task is
$0.708$\,K (bamboo panel, whose conductivity lies near the
lower boundary of the sampled range, where operator error is
largest); this is comparable to the
bamboo--adobe gap itself, and the FDM values are therefore
retained as the ranking basis; the PINO ranking order
nevertheless agrees with FDM for all five materials.

\subsection{Material Ranking Across the Climate Design Space}
\label{sec:climate_sweep}

The ranking above is reported at a single nominal Sindh climate
condition. Because the daily-maximum outdoor temperature
$T_{\mathrm{out,max}}$ and
peak solar irradiance $G_{s,\mathrm{peak}}$ are themselves swept parameters within
$\boldsymbol{\mu}$ \eqref{eq:params} (Table~\ref{tab:params}),
the trained PINO $\mathcal{G}_\theta$ can be queried across the
full $(T_{\mathrm{out,max}}, G_{s,\mathrm{peak}})$ design space without retraining,
enabling the material ranking to be characterised as a function
of climate rather than at a single operating point.

To map the full ranking surface, the trained operator is queried across a $20 \times 20$ climate grid.
$J(\boldsymbol{\mu})$ \eqref{eq:qoi_peak} is evaluated uniformly
over
$T_{\mathrm{out,max}} \in [\SI{25}{\celsius}, \SI{48}{\celsius}]$
and
$G_{s,\mathrm{peak}} \in [\SI{200}{\watt\per\metre\squared}, \SI{900}{\watt\per\metre\squared}]$
(the full ranges of Table~\ref{tab:params}), holding
$T_{\mathrm{in}} = \SI{35}{\celsius}$, $w_0 = 0$, and
$L = \SI{0.25}{\metre}$ fixed at their nominal values, for all
five candidate materials (Table~\ref{tab:materials}). This
gives 400 PINO evaluations per material (2000 in total),
each obtained in approximately 3~ms per sample on a commodity
GPU (Tesla T4) using the already-trained operator, requiring
no additional FDM data generation or retraining.

To confirm the predicted ranking structure against ground
truth, 9 climate points spanning the corners, edge midpoints,
and centre of the swept region are additionally solved with the
FDM solver for all five materials (45 FDM solutions in total);
Table~\ref{tab:spot} reports the resulting peak temperatures
and winners.
The FDM spot checks confirm the regime structure predicted by
the operator: at sub-ambient outdoor conditions with low to
moderate solar gain
($T_{\mathrm{out,max}} = \SI{25}{\celsius}$ with
$G_{s,\mathrm{peak}} \le \SI{550}{\watt\per\metre\squared}$),
fired clay brick achieves the lowest peak temperature among
all candidates, while at every heating-dominated point the
bamboo panel wins overall and clay--straw adobe wins among
widely available materials, in agreement with the
PINO-predicted winner map.

\begin{table}[htbp]
\centering
\caption{FDM ground-truth spot checks of the climate sweep:
         peak inner-surface temperature $J$ (\si{\celsius}) on
         the final periodic day for all five materials at 9
         climate points ($T_{\mathrm{in}} = \SI{35}{\celsius}$,
         $w_0 = 0$, $L = \SI{0.25}{\metre}$). Bold marks the
         winner among all candidates; the winner among
         widely available (\textbf{H}) materials is fired clay
         brick at the first two rows and clay--straw adobe at
         all others.}
\label{tab:spot}
\begin{tabular}{ccccccc}
\toprule
$T_{\mathrm{out,max}}$ (\si{\celsius})
    & $G_{s,\mathrm{peak}}$ (\si{\watt\per\metre\squared})
    & Bamboo & Adobe & Mud & Lime--mud & Fired \\
\midrule
25.0 & 200 & 33.55 & 33.22 & 32.92 & 32.69 & \textbf{32.39} \\
25.0 & 550 & 33.97 & 33.85 & 33.74 & 33.73 & \textbf{33.55} \\
25.0 & 900 & \textbf{34.39} & 34.50 & 34.60 & 34.81 & 34.76 \\
36.5 & 200 & \textbf{34.83} & 34.91 & 34.98 & 35.09 & 35.08 \\
36.5 & 550 & \textbf{35.25} & 35.54 & 35.80 & 36.13 & 36.25 \\
36.5 & 900 & \textbf{35.67} & 36.19 & 36.66 & 37.21 & 37.46 \\
48.0 & 200 & \textbf{36.11} & 36.60 & 37.04 & 37.48 & 37.78 \\
48.0 & 550 & \textbf{36.52} & 37.23 & 37.87 & 38.52 & 38.94 \\
48.0 & 900 & \textbf{36.95} & 37.88 & 38.72 & 39.60 & 40.15 \\
\bottomrule
\end{tabular}
\end{table}

\begin{figure}[htbp]
\centering
\includegraphics[width=\textwidth]{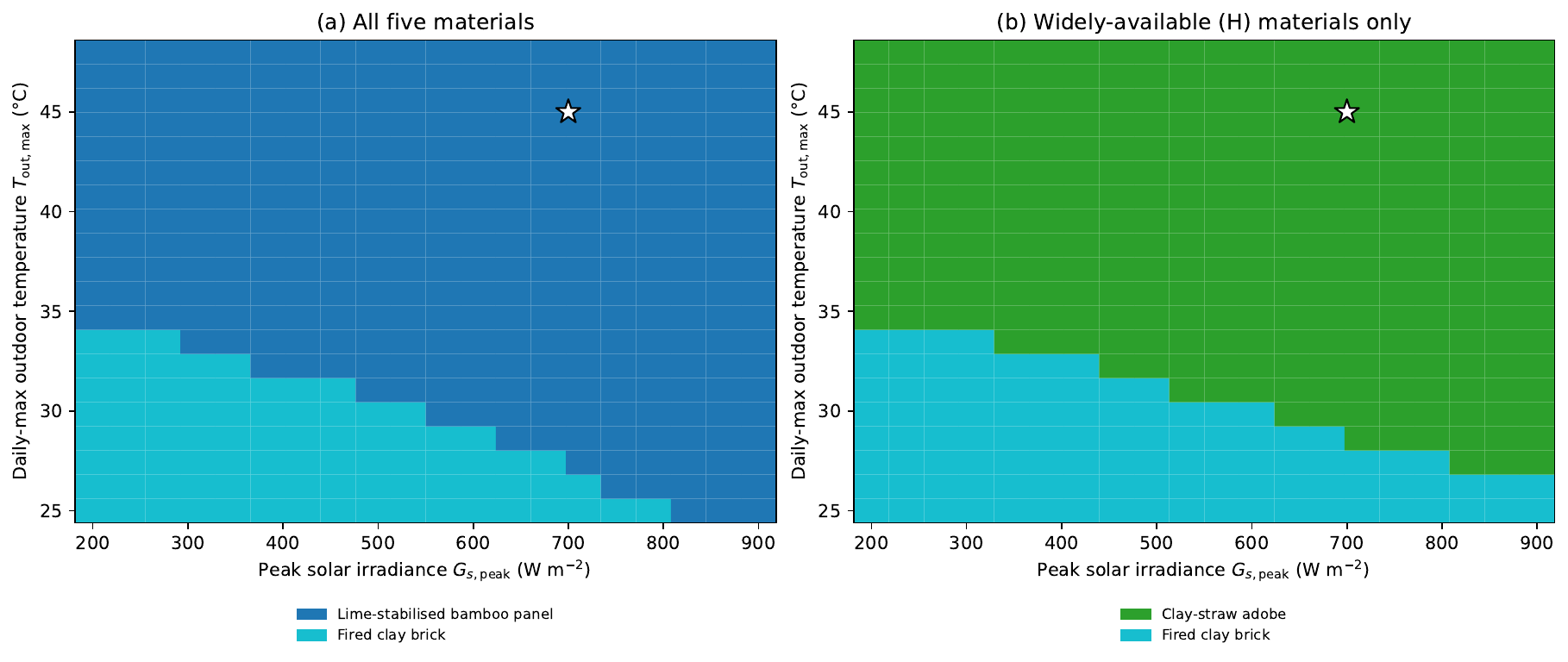}
\caption{Winning material (lowest $J(\boldsymbol{\mu})$) as a
         function of daily-maximum outdoor temperature
         $T_{\mathrm{out,max}}$ and peak solar irradiance
         $G_{s,\mathrm{peak}}$, evaluated via the trained PINO
         $\mathcal{G}_\theta$ across a
         $20 \times 20$ grid spanning the full ranges of
         Table~\ref{tab:params}
         ($T_{\mathrm{in}} = \SI{35}{\celsius}$, $w_0 = 0$,
         $L = \SI{0.25}{\metre}$).
         (a)~Winning material among all five candidates.
         (b)~Winning material restricted to widely available
         (\textbf{H}) materials only. The marker indicates the
         single nominal condition
         ($T_{\mathrm{out,max}} = \SI{45}{\celsius}$,
          $G_{s,\mathrm{peak}} = \SI{700}{\watt\per\metre\squared}$)
         used in Section~\ref{sec:ranking}.}
\label{fig:climate_heatmap}
\end{figure}

The operator reveals a clear thermal regime boundary separating two distinct material-selection strategies.
Figure~\ref{fig:climate_heatmap} presents the resulting material
ranking as a function of $(T_{\mathrm{out,max}}, G_{s,\mathrm{peak}})$. Panel (a)
shows the winning material among all five candidates at each
grid point; panel (b) restricts the comparison to the three
widely available (\textbf{H}) materials (clay--straw adobe,
mud brick, fired clay brick), reflecting the practical
recommendation set used in Section~\ref{sec:cost}.

\begin{figure}[htbp]
\centering
\includegraphics[width=0.65\textwidth]{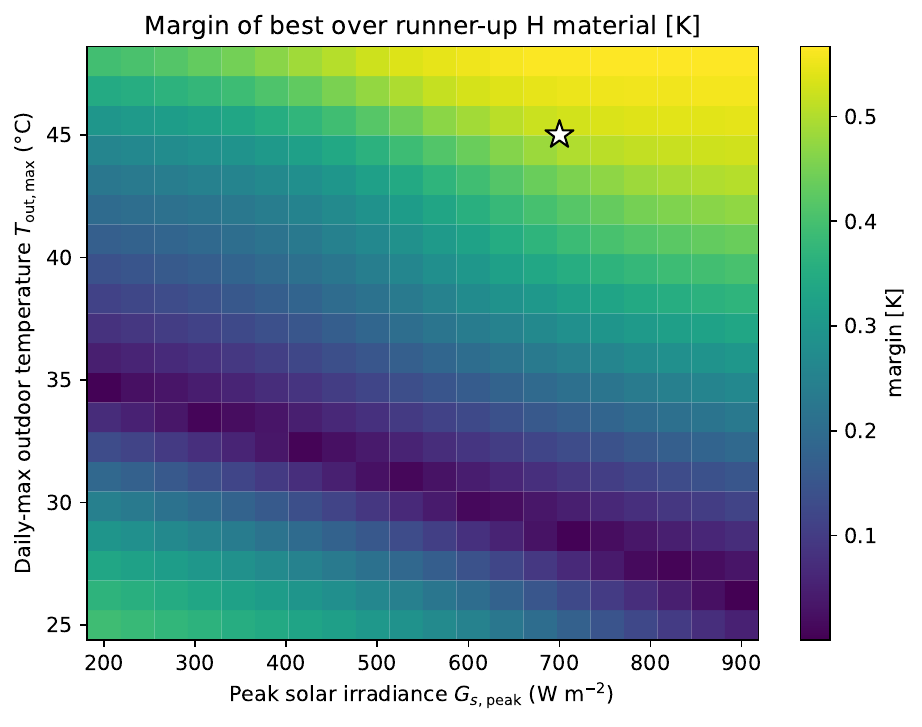}
\caption{Recommendation robustness: margin (K) between the
         best and runner-up widely available (\textbf{H})
         material at each point of the climate grid, evaluated
         via the trained PINO. The margin vanishes along the
         heat-exclusion/heat-rejection regime boundary of
         Section~\ref{sec:climate_sweep_limitation}.}
\label{fig:climate_margin}
\end{figure}

According to the operator's winner map, lime-stabilised bamboo
panel wins on 79\% of the grid among all five candidates and
fired clay brick on the remaining 21\%, concentrated at
sub-ambient outdoor conditions; among widely available
materials, clay--straw adobe wins on 74\% of the grid and
fired clay brick on 26\% (panel b). Mud brick and lime--mud
composite win at no point of the swept domain and therefore do
not appear in either panel: both are dominated across the entire
climate design space by lower-diffusivity alternatives, so the
effective competition reduces to three materials.
Figure~\ref{fig:climate_margin} complements the winner map
with the \emph{margin} by which the best widely available
material beats the runner-up at each grid point: the median
margin is \SI{0.24}{\kelvin} and the maximum
\SI{0.57}{\kelvin}, decaying to zero along the boundary
between the two regimes, so that the recommendation is most
robust deep inside each regime and genuinely indifferent near
the boundary.

\subsubsection{Ranking Regimes: Heat Exclusion versus Heat
Rejection}
\label{sec:climate_sweep_limitation}

The fired-brick region identified by the operator and confirmed
by the FDM spot checks (Table~\ref{tab:spot}) corresponds to a
qualitatively distinct and physically meaningful thermal
regime.
With the indoor air held at $T_{\mathrm{in}} = \SI{35}{\celsius}$,
sub-ambient outdoor conditions with low solar gain drive the
daily-mean sol-air temperature \eqref{eq:solair} below the
indoor temperature, and the envelope's function inverts: rather
than \emph{excluding} external heat, the wall's task is to
\emph{reject} indoor heat to the cooler exterior, and the most
conductive material (fired clay brick,
$k_{\mathrm{dry}} = \SI{0.65}{\watt\per\metre\per\kelvin}$)
then minimises the peak inner-surface temperature.
Conversely, throughout the heating-dominated majority of the
design space, which encompasses the daytime summer
conditions relevant to rural Sindh, the insulating,
high-thermal-mass materials win, and the nominal-condition
ranking of Table~\ref{tab:ranking} applies.
The recommendation of clay--straw adobe among widely available
materials (Section~\ref{sec:cost}) is therefore robust across
heating-dominated conditions, while the regime boundary itself,
resolved consistently by the operator and by FDM ground
truth, delineates where a different selection logic applies,
for example for structures that are internally heated by
occupancy or cooking loads under mild outdoor conditions.

\subsection{Global Sensitivity Analysis}
\label{sec:sobol}

\begin{figure}[htbp]
\centering
\includegraphics[width=0.7\textwidth]{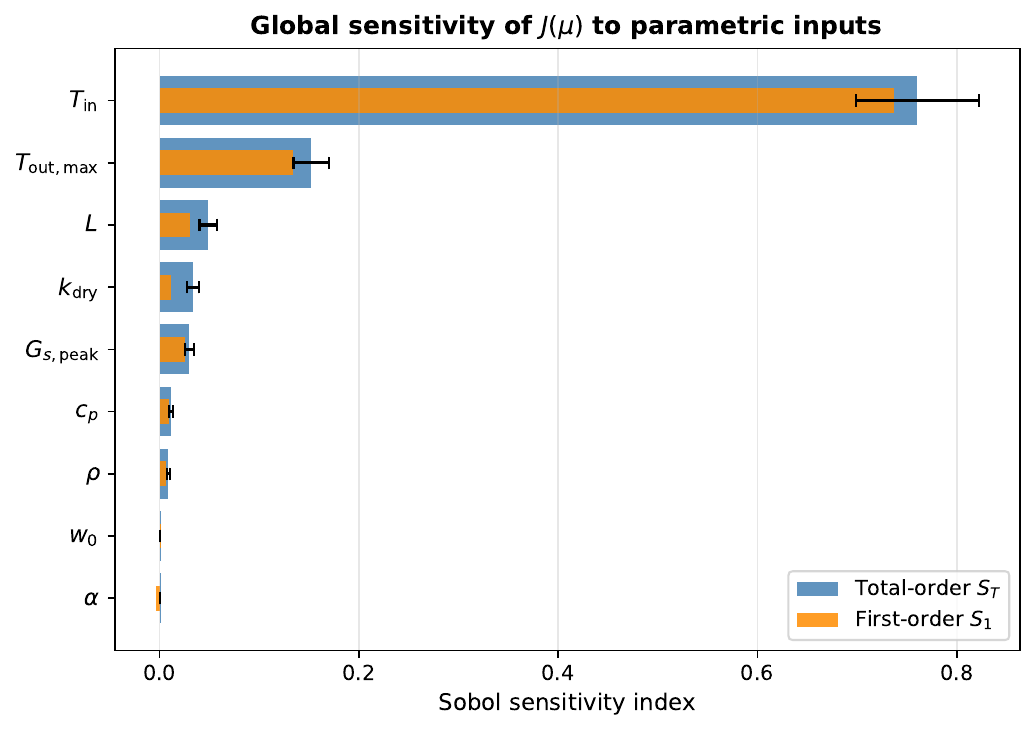}
\caption{First-order ($S_1$) and total-order ($S_T$) Sobol'
         sensitivity indices for the nine parameters of
         $\boldsymbol{\mu}$, ranked by influence on the peak
         inner surface temperature $J(\boldsymbol{\mu})$.}
\label{fig:sobol}
\end{figure}

To identify which entries of the nine-dimensional parameter
vector $\boldsymbol{\mu}$ \eqref{eq:params} most strongly govern
the peak inner surface temperature $J(\boldsymbol{\mu})$
\eqref{eq:qoi_peak} across the full parametric range
(Table~\ref{tab:params}), a variance-based global sensitivity
analysis is performed using first-order ($S_1$) and total-order
($S_T$) Sobol' indices~\cite{Saltelli2010}, estimated via
Saltelli sampling (base sample size $N=1024$,
\texttt{calc\_second\_order=False}, giving $N(D+2)=11{,}264$
evaluations for $D=9$ parameters) using the trained PINO
$\mathcal{G}_\theta$ as a computationally efficient surrogate for
the FDM solver. Table~\ref{tab:sobol} reports the resulting
indices for all nine parameters, sorted by descending $S_T$.
Figure~\ref{fig:sobol} presents the same results graphically.

\begin{table}[htbp]
\centering
\caption{First-order ($S_1$) and total-order ($S_T$) Sobol'
         sensitivity indices of the nine parameters in
         $\boldsymbol{\mu}$ \eqref{eq:params} with respect to
         the peak inner surface temperature
         $J(\boldsymbol{\mu})$ \eqref{eq:qoi_peak}, sorted by
         descending $S_T$, with 95\% confidence intervals from
         Jansen's estimator, per the recommendation of~\cite{Saltelli2010}. Estimated via Saltelli sampling
         (11,264 evaluations) using the trained PINO as a
         surrogate. Slightly negative $S_1$ values are
         Monte-Carlo noise, consistent with zero within the
         confidence intervals.}
\label{tab:sobol}
\begin{tabular}{clcc}
\toprule
Rank & Parameter & $S_1$ & $S_T$ \\
\midrule
1 & $T_{\mathrm{in}}$          & $0.737 \pm 0.071$  & $0.760 \pm 0.061$ \\
2 & $T_{\mathrm{out,max}}$     & $0.134 \pm 0.035$  & $0.152 \pm 0.018$ \\
3 & $L$                        & $0.031 \pm 0.019$  & $0.049 \pm 0.009$ \\
4 & $k_{\mathrm{dry}}$         & $0.011 \pm 0.017$  & $0.034 \pm 0.006$ \\
5 & $G_{s,\mathrm{peak}}$      & $0.026 \pm 0.014$  & $0.030 \pm 0.004$ \\
6 & $c_p$                      & $0.009 \pm 0.008$  & $0.011 \pm 0.002$ \\
7 & $\rho$                     & $0.007 \pm 0.008$  & $0.009 \pm 0.002$ \\
8 & $w_0$                      & $0.000 \pm 0.002$  & $0.001 \pm 0.000$ \\
9 & $\alpha$                   & $-0.003 \pm 0.002$ & $0.000 \pm 0.000$ \\
\bottomrule
\end{tabular}
\end{table}

Indoor air temperature $T_{\mathrm{in}}$ dominates the global
variance of $J(\boldsymbol{\mu})$ ($S_T = 0.760$), followed by
the daily-maximum outdoor temperature $T_{\mathrm{out,max}}$
($S_T = 0.152$). This is physically consistent with the
governing Robin boundary condition at the inner surface
(Eq.~\eqref{eq:bc_in}), where $T_{\mathrm{in}}$ enters as a
direct additive offset at the exact location where $J$ is
evaluated, and it is reinforced by the strong attenuation
quantified by the decrement factors of
Table~\ref{tab:ranking} ($f = 0.016$--$0.090$): because the
envelope suppresses the external thermal wave by one to two
orders of magnitude, the inner surface is anchored to the
indoor air temperature, and the \SI{20}{\celsius} sweep of
$T_{\mathrm{in}}$ necessarily explains most of the variance in
the absolute surface temperature, itself evidence that the
candidate envelopes perform their attenuating function.
Wall thickness ($S_T = 0.049$) and the four material properties
($k_{\mathrm{dry}}$, $c_p$, $\rho$, $\alpha$; individual $S_T$
values summing to $\approx 0.05$, a rough additive approximation
since total-order indices are not strictly additive under
parameter interactions) form a second tier of comparable magnitude,
with the solar peak $G_{s,\mathrm{peak}}$ ($S_T = 0.030$)
alongside. Moisture-related parameters ($w_0$, $\alpha$) show
negligible sensitivity ($S_T \leq 0.001$), consistent with the
static moisture treatment over the simulation window
(Section~\ref{sec:ics}).

This global ranking should be interpreted alongside, rather
than in place of, the material ranking in
Section~\ref{sec:ranking}. The two analyses answer different
questions: the Sobol' indices quantify sensitivity across the
\emph{entire} nine-dimensional design space, in which
$T_{\mathrm{in}}$, $T_{\mathrm{out,max}}$, and $L$ vary
freely, whereas the material ranking in
Table~\ref{tab:ranking} is performed at \emph{fixed} climate
and geometry, isolating material properties as the only source
of variation between candidates. Conditional on climate and
geometry, material properties are therefore the governing
drivers of \emph{relative} material performance (including
the dynamic metrics $\varphi$ and $f$, which are almost
entirely material- and thickness-controlled), even though
they account for a small share of the total variance in the
\emph{absolute} peak temperature when climate is also allowed
to vary. This is consistent with the climate-sweep result in
Section~\ref{sec:climate_sweep}: within each regime, material
properties determine which candidate performs best, while the
climate variables set the overall temperature level and, at the
regime boundary, decide \emph{which selection logic}
(heat exclusion or heat rejection) applies.
\subsection{Material Cost--Performance Assessment}
\label{sec:cost}

Identifying the thermally optimal material alone does not
satisfy the study objective: the final recommendation must
be cost-effective and practically accessible to rural Sindh
communities.
This requirement follows established practice in comparable
low-income, hot-climate housing research: a life-cycle cost
study of mud brick and earthbag walls in rural Uganda jointly
evaluated thermal performance and life-cycle cost rather than
either factor in isolation, finding that the earthbag system, whose much lower assembly
U-value reflects its far greater wall thickness as well as
its material properties, also achieved a lower total
life-cycle cost than the conventional
alternative~\cite{Wesonga2023}.
That study's life-cycle cost model includes the discounted
cost of mechanical cooling energy, computed from cooling
degree-days, over a 30-year horizon, which is appropriate
for conditioned housing in its tropical setting; the
present CPI instead targets passively cooled, unconditioned
rural Sindh housing, where the relevant quantity is the
reduction in peak indoor surface temperature per unit
construction cost rather than a discounted energy bill.
This section defines a cost--performance index (CPI) that
combines the simulated thermal performance with indicative
material cost and local availability.

\subsubsection{Reference Conditions and Cost Data}
\label{sec:cost_data}

All costs are evaluated at the reference wall thickness
$L_{\mathrm{ref}} = \SI{0.25}{\metre}$ (midpoint of the
parametric range, Table~\ref{tab:params}).
Indicative material costs $C_{\mathrm{mat}}$ (PKR/m$^3$)
for rural Sindh are listed in Table~\ref{tab:cost};
these are authors' estimates based on local market enquiries
in rural Sindh (2025--2026).
These are approximate 2025--2026 market values excluding
skilled labour and transport; they should be confirmed
from local supplier records before final reporting.
The steady-state thermal resistance at $L_{\mathrm{ref}}$
is $R_i = L_{\mathrm{ref}}/k_{\mathrm{dry},i}$
(\si{\metre\squared\kelvin\per\watt}), independent of the
PINO results and computable directly from Table~\ref{tab:materials}.

\begin{table}[htbp]
\centering
\caption{Cost and cost--performance indices for the five
         candidate materials at reference thickness
         $L_{\mathrm{ref}} = 0.25$\,m.
         $C_{\mathrm{mat}}$: approximate market cost of raw
         material (PKR/m$^3$, rural Sindh, 2025--2026,
         excluding labour) (authors' market estimates).
         $R_i$: steady-state thermal resistance at $L_{\mathrm{ref}}$.
         CPI$_{\mathrm{static}}$: normalised $R_i / C_i$
         (see \eqref{eq:cpi_static}).
         CPI$_{\mathrm{dyn}}$: normalised $(J_{\mathrm{max}}-J_i)/C_i$
         (see \eqref{eq:cpi_dyn}; $J_i$ from Table~\ref{tab:ranking}).
         Both CPIs: 1.00 = best cost--performance.
         CPI$_{\mathrm{dyn}} = 0.000$ for fired clay brick is a normalisation artefact (worst-performing material, zero benefit relative to itself).
         Availability: \textbf{H}\,=\,widely available in rural
         Sindh; \textbf{M}\,=\,regionally available;
         \textbf{L}\,=\,limited/emerging.
         Rows sorted by CPI$_{\mathrm{static}}$ (best first).}
\label{tab:cost}
\begin{tabular}{lrrrrcc}
\toprule
Material
    & $C_{\mathrm{mat}}$
    & Cost/m$^2$
    & $R_i$
    & CPI$_{\mathrm{static}}$
    & CPI$_{\mathrm{dyn}}$
    & Avail.\ \\
    & (PKR/m$^3$)
    & (PKR)
    & (m$^2$K/W)
    & (norm.)
    & (norm.)
    & \\
\midrule
Clay--straw adobe            & 3\,500  & 875    & 0.71 & 1.00 & 1.000 & \textbf{H} \\
Mud brick (unfired)          & 4\,500  & 1\,125 & 0.56 & 0.61 & 0.490 & \textbf{H} \\
Lime--mud composite          & 8\,000  & 2\,000 & 0.45 & 0.28 & 0.104 & \textbf{M} \\
Lime-stabilised bamboo panel & 20\,000 & 5\,000 & 1.00 & 0.24 & 0.248 & \textbf{L} \\
Fired clay brick             & 10\,500 & 2\,625 & 0.38 & 0.18 & 0.000 & \textbf{H} \\
\bottomrule
\end{tabular}
\end{table}

\subsubsection{Cost--Performance Index}
\label{sec:cpi}

Multi-criteria composite indices combining thermal performance
with cost and other sustainability attributes represent an
established approach to building material selection~\cite{Akadiri2013};
the present study adopts a simplified ratio-based formulation
that is directly computable from simulation output and
indicative material costs, without requiring expert elicitation
of criterion weights.

\noindent\textbf{Static CPI.}
The static index uses the steady-state thermal resistance as
the thermal metric and is available before PINO training:
\begin{equation}
    \mathrm{CPI}_{\mathrm{static},i}
    \;=\;
    \frac{R_i}{C_i}
    \;=\;
    \frac{1}{k_{\mathrm{dry},i}\cdot C_{\mathrm{mat},i}}
    \label{eq:cpi_static}
\end{equation}
where $C_i = C_{\mathrm{mat},i}\cdot L_{\mathrm{ref}}$
is the wall cost per unit area (PKR/m$^2$).
The normalised values in Table~\ref{tab:cost} are obtained
by dividing each $\mathrm{CPI}_{\mathrm{static},i}$ by the
maximum over all five materials.
Clay--straw adobe achieves the highest static CPI (1.00),
combining the lowest raw cost with moderate thermal resistance.
Fired clay brick scores worst (0.18) despite its wide local
availability, because its high conductivity and firing cost
jointly penalise it.
Lime-stabilised bamboo panel has the highest absolute
$R$-value (1.00\,m$^2$K/W) but poor CPI (0.24) due to
its high estimated cost.

\medskip\noindent\textbf{Dynamic CPI.}
The static index does not capture thermal mass.
The dynamic index uses the simulated peak inner surface
temperature $J_i = J(\boldsymbol{\mu}_i^{\mathrm{nom}})$
\eqref{eq:qoi_peak} from the FDM ground-truth results in
Table~\ref{tab:ranking}:
\begin{equation}
    \mathrm{CPI}_{\mathrm{dyn},i}
    \;=\;
    \frac{J_{\mathrm{max}} - J_i}{C_i}
    \quad [\text{K per PKR/m}^2]
    \label{eq:cpi_dyn}
\end{equation}
where $J_{\mathrm{max}} = \max_j J_j = \SI{38.76}{\celsius}$
(fired clay brick, the worst-performing material at the nominal
heating-dominated condition) and
$C_i = C_{\mathrm{mat},i} \cdot L_{\mathrm{ref}}$ is the
wall cost per unit area at $L_{\mathrm{ref}} = \SI{0.25}{\metre}$.
Normalised values (dividing by the maximum $\mathrm{CPI}_{\mathrm{dyn}}$)
are added to Table~\ref{tab:cost}.
$\mathrm{CPI}_{\mathrm{dyn}}$ gives the cooling benefit in kelvin
per PKR/m$^2$ of wall cost; a higher value means more cooling
per rupee spent.

\medskip\noindent\textbf{Final recommendation criterion.}
The recommended material for rural Sindh construction is that
which simultaneously achieves a high
$\mathrm{CPI}_{\mathrm{dyn}}$ and an availability rating of
\textbf{H}, ensuring the thermal advantage can be realised
without supply-chain risk for remote communities.
\section{Discussion}
\label{sec:discussion}

The modest improvement of PINO over the data-only FNO baseline
on the temperature field (5.1\% reduction in relative $L^2$
error at $N_{\mathrm{train}} = 1200$) is expected given the
nature of the problem: the governing PDE \eqref{eq:heat} is a
one-dimensional linear diffusion equation on a uniform grid,
which is among the smoothest and most data-amenable problems
for operator learning~\cite{Li2021FNO}.
In this regime, a well-trained FNO already captures the
solution operator to high accuracy from 1200 samples alone,
and the aggregate QoI accuracy of the two models is
correspondingly close (MAE on $J$ of 0.201~K vs.\ 0.205~K).
The value of the physics loss appears instead in two specific
places.
First, it is spatially targeted: the error-versus-position
profile (Fig.~\ref{fig:err_position}) shows a 15.5\% RMSE
reduction at the inner surface $\xi = 1$, precisely the
location where the QoI is evaluated and where accuracy matters
most for material ranking.
Second, it substitutes for data: in the data-scarce regime
(Section~\ref{sec:data_eff}), PINO reduces the QoI MAE by
18.8\% at $N_{\mathrm{train}} = 150$ and 27.5\% at
$N_{\mathrm{train}} = 300$, and PINO trained on 150 FDM
samples matches the data-only FNO trained on twice as many,
directly halving the high-fidelity data budget at that accuracy
level; this is the practically relevant benefit when each
additional sample requires a full transient simulation.

The material ranking produced by PINO is identical to the FDM
ground truth across all five candidates.
The maximum prediction error on the ranking task is 0.708~K
(bamboo panel, Table~\ref{tab:ranking}), comparable to the
bamboo--adobe gap itself; the FDM values are therefore retained
as the basis of all downstream tables, with the operator
serving as the rapid-evaluation tool whose ranking agreement is
verified rather than assumed.
The thermal ranking is physically interpretable and now
three-dimensional: lime-stabilised bamboo panel ranks first on
peak temperature, time lag, and decrement factor alike, due to
its combination of low conductivity
($k_{\mathrm{dry}} = \SI{0.25}{\watt\per\metre\per\kelvin}$)
and high specific heat capacity
($c_p = \SI{1800}{\joule\per\kilogram\per\kelvin}$), which
together minimise thermal diffusivity and thus maximise both
attenuation and delay of the diurnal wave; fired clay brick
ranks last on all three metrics under heating-dominated
conditions because its high conductivity
($k_{\mathrm{dry}} = \SI{0.65}{\watt\per\metre\per\kelvin}$)
dominates despite its moderate density.
Under cyclic loading the inter-material gaps in peak
temperature are compressed relative to what steady resistance
alone would suggest, while the dynamic metrics
(Section~\ref{sec:dyn_metrics}), with a 13.4~h versus 7.6~h
time lag and a factor-of-six spread in decrement, differentiate
the candidates sharply, demonstrating
the value of the periodic-day formulation for envelope
comparison.

Beyond the single nominal condition, the climate sweep in
Section~\ref{sec:climate_sweep} shows this ranking is stable
across the heating-dominated majority of the swept
$(T_{\mathrm{out,max}}, G_{s,\mathrm{peak}})$ design space,
and it resolves, in agreement with 45 FDM ground-truth
solutions, a physically meaningful regime boundary at
sub-ambient outdoor conditions where the envelope's function
inverts from heat exclusion to heat rejection and conductive
fired clay brick becomes optimal
(Section~\ref{sec:climate_sweep_limitation}); the margin map
(Fig.~\ref{fig:climate_margin}) quantifies how robust the
recommendation is at each climate point. The global
sensitivity analysis in Section~\ref{sec:sobol} further shows
that, while indoor temperature dominates the absolute
temperature level across the full parametric range (a
consequence of the strong attenuation the envelopes provide),
material properties remain the governing drivers of relative
material performance at fixed climate and geometry, reconciling the Sobol' ranking with the material ranking reported above.
Beyond the specific Sindh application, two implications carry
to the broader community.
For building-energy practitioners, the study demonstrates a
transferable workflow: a validated, inexpensive one-dimensional
solver paired with a physics-informed operator turns exhaustive
design-space exploration (400-point climate maps,
$\sim$$10^4$-evaluation sensitivity analyses) into
interactive-speed computation, a pattern that applies wherever
envelope options must be screened across wide climatic and
material ranges, well beyond the present case study.
Unlike prior material comparison studies for the Sindh region
that evaluate discrete material configurations at fixed
parameter combinations~\cite{Ali2025}, the trained PINO
enables continuous evaluation across the full
nine-dimensional parameter space without re-running the
FDM solver for each new combination.
For the scientific machine-learning community, the results
delineate \emph{when} PDE-constrained operator learning pays
off for smooth parabolic problems: not in aggregate accuracy at
generous data volumes, but in data-scarce regimes and at
physically critical boundaries, guidance that is directly
actionable whenever each additional training sample requires an
expensive simulation or experiment.

Several limitations of the present study should be noted.
The physical model is one-dimensional and assumes spatially
uniform, time-invariant moisture content and no latent heat of
vaporisation; these choices are justified by the moisture
timescale argument in Section~\ref{sec:ics} but are
approximations nonetheless.
The diurnal forcing model fixes the shape of the daily cycle: a
clear-sky half-sine irradiance profile, a sinusoidal air
temperature with a fixed \SI{12}{\kelvin} swing, identical
consecutive days, and a constant indoor air temperature;
day-to-day weather variability, cloud transients, and coupled
indoor-air dynamics are outside the present scope.
The moisture--conductivity coefficients $\alpha$ are
author-assigned within literature-informed ranges rather than
experimentally measured; the same applies to the base thermal
properties ($k_{\mathrm{dry}}$, $\rho$, $c_p$) of the five
candidate materials, which are drawn from published values for
comparable earthen and bio-based composites rather than measured
on physical specimens. Any residual uncertainty in these values
is, however, immaterial to the ranking QoI, since the global
sensitivity analysis attributes a combined total-order index of
only $\approx 0.05$ to the four material properties, and
$S_T \approx 0$ to $\alpha$, across their entire sampled ranges
(Table~\ref{tab:sobol}); experimental characterisation of the
candidate materials is identified as future work.
The cost data in Table~\ref{tab:cost} are indicative 2025--2026
market estimates for rural Sindh and should be verified against
current local supplier records before use in procurement
decisions.
No experimental validation against physical wall specimens has
been performed; thermocouple measurements in constructed
specimens of the five candidate materials would confirm whether
the FDM thermal properties in Table~\ref{tab:materials}
are representative of materials sourced from rural Sindh
suppliers.
These limitations are consistent with the scope of a
parametric evaluation study and are addressed in the future
work directions of Section~\ref{sec:conclusion}.

\section{Conclusions}
\label{sec:conclusion}

This study presented a two-stage computational framework for
parametric thermal analysis of five indigenous wall materials
for rural Sindh housing, motivated by the urgent need for
evidence-based material selection in post-flood reconstruction
programmes.
The Crank--Nicolson FDM solver, upgraded with diurnal
time-varying boundary forcing and a periodic quasi-steady
extraction protocol, achieves second-order convergence in the
time-varying setting (\SI{0.58}{\milli\kelvin} inner-surface
error at production resolution), passes the Robin zero-drift
test to within $10^{-12}$~K, and reaches a verified periodic
state on every one of the 1500 dataset samples.
The trained PINO attains a relative $L^2$ field error of
$5.14\times10^{-4}$ and a 0.201~K MAE on the peak inner
surface temperature, reproducing the FDM material ranking
exactly; a data-efficiency study shows the physics loss is most
valuable when high-fidelity data are scarce, reducing the QoI
error by 18.8--27.5\% at 150--300 training samples and
allowing PINO trained on 150 samples to match a data-only FNO
trained on 300, halving the FDM data budget, while the
error-versus-position analysis localises the physics benefit at
the inner surface (15.5\% RMSE reduction) where the QoI is
evaluated.
The periodic-day formulation additionally yields the ISO~13786
dynamic metrics, which the operator reproduces to
\SI{0.99}{\hour} (time lag) and 0.010 (decrement factor) MAE.
Among the five candidate materials, lime-stabilised bamboo
panel achieves the lowest peak inner surface temperature
($J_{\mathrm{FDM}} = \SI{36.37}{\celsius}$), the longest time
lag (\SI{13.4}{\hour}), and the strongest attenuation
($f = 0.016$), but carries limited local availability
(\textbf{L}) and the highest material cost
(PKR~20{,}000/m$^3$).
Among widely available materials (\textbf{H}), clay--straw
adobe is the recommended choice for rural Sindh construction:
it achieves the best dynamic cost--performance index
(CPI$_{\mathrm{dyn}} = 1.00$), a peak inner surface
temperature only 0.70~K above the bamboo panel, a
\SI{10.1}{\hour} time lag that displaces the indoor heat peak
into the late evening, and a material cost of
PKR~3{,}500/m$^3$, the lowest of all five candidates.
A systematic sweep of the trained operator across the full
climate design space (Section~\ref{sec:climate_sweep}),
confirmed by 45 FDM ground-truth solutions, showed that this
ranking is stable across heating-dominated conditions and
resolved a physically meaningful regime boundary at sub-ambient
outdoor conditions, where the envelope's function inverts from
heat exclusion to heat rejection and conductive fired clay
brick becomes optimal
(Section~\ref{sec:climate_sweep_limitation}).
A global sensitivity analysis (Section~\ref{sec:sobol}) further
identified indoor air temperature as the dominant driver of the
absolute peak inner surface temperature (a direct
consequence of the strong attenuation the candidate envelopes
provide), while confirming that material properties remain
the primary determinant of relative material ranking at fixed
climate and geometry.

\medskip
\noindent
\textbf{Future work.}
Natural extensions include:
(i) making $D_w$ moisture-dependent for higher physical
fidelity~\cite{Wei2019};
(ii) replacing the idealised periodic day with measured
multi-day weather sequences (e.g.\ NASA POWER hourly files),
including cloud transients and day-to-day variability;
(iii) experimental validation against thermocouple
measurements in physical wall specimens of the five
candidate materials;
(iv) extension to multi-day peak-summer simulations using a
fully coupled heat--moisture PDE system with
vapour-pressure-driven transport and latent heat of
vaporisation, consistent with the hot-dry low-humidity climate
of rural Sindh, for which
$\tau_m \sim 26$~days (Eq.~\eqref{eq:tau_m}) renders moisture
redistribution dynamically significant over the simulation
window;
and (v) replacing the manually tuned physics-loss weight
$\lambda_T$ with adaptive or self-calibrating weighting
schemes, removing the pilot grid search from the workflow.

\section*{CRediT Authorship Contribution Statement}

\textbf{Muhammad Akbar Khan:}
Conceptualization,
Methodology,
Software,
Formal analysis,
Investigation,
Data curation,
Visualization,
Writing -- original draft.
\textbf{Fahim Raees:}
Conceptualization,
Methodology,
Supervision,
Writing -- review \& editing.
\textbf{Ubaida Fatima:}
Formal analysis,
Validation,
Writing -- review \& editing.

\section*{Declaration of Competing Interest}

The authors declare that they have no known competing financial
interests or personal relationships that could have appeared to
influence the work reported in this paper.

\section*{Funding}

This work was supported by the Sindh Higher Education Commission,
Government of Sindh, under the Sindh Research Support Program
(grant no.\ SRSP-321), and by NED University of Engineering \&
Technology, Karachi, Pakistan.

\section*{Acknowledgements}

The authors thank the Department of Mathematics, NED University
of Engineering \& Technology, for computational resources and
administrative support.

\section*{Data Availability}

The FDM solver, PINO training code, the 1500-sample LHS dataset,
and the NASA POWER climate forcing verification package generated
in this study are openly available on
Zenodo~\cite{Khan2026Zenodo}; the DOI
\url{https://doi.org/10.5281/zenodo.21311299} resolves to the
latest version.

\bibliographystyle{unsrt}

\end{document}